\documentclass[12pt,a4paper]{article}

\usepackage[british]{babel}
\usepackage[T1]{fontenc}
\usepackage{lmodern}
\usepackage{microtype}

\usepackage[a4paper,top=2cm,bottom=2cm,left=2.5cm,right=2.5cm]{geometry}

\usepackage{amsmath}
\usepackage{amsfonts}
\usepackage{mathrsfs}
\usepackage{graphicx}
\usepackage{subcaption}
\usepackage{booktabs}
\usepackage{threeparttable}
\usepackage{diagbox}

\usepackage[hidelinks]{hyperref}
\usepackage{url}

\usepackage{algorithm}
\usepackage{algorithmicx}
\usepackage{algpseudocode}
\usepackage{listings}

\usepackage{setspace}
\usepackage{titlesec}
\titleformat{\section}
  {\normalfont\Large\bfseries}{\thesection.}{1em}{}

\usepackage{caption}
\usepackage[authoryear,round]{natbib}
\date{}
\begin{document}
\begingroup
\renewcommand{\thefootnote}{\fnsymbol{footnote}}

\begin{center}
{\LARGE Moonworks Lunara Aesthetic II: An Image Variation Dataset} \\[4.2ex]

{\small Yan Wang, Partho Hassan, Samiha Sadeka\footnote{Griffith University,  s.sadeka@griffithuni.edu.au}, \\ Nada Soliman\footnote{Carnegie Mellon University,  nsoliman@alumni.cmu.edu}, Sayeef Abdullah and Sabit Hassan} \\ [1.5em]

{\small \texttt{research@moonworks.ai}}\\ [1.5em]

\end{center}
\endgroup
\setcounter{footnote}{0}

\begin{figure}[!h]
    \centering
    \setlength{\tabcolsep}{2pt}
    \begin{tabular}{cccccc}
        \includegraphics[width=0.16\textwidth]{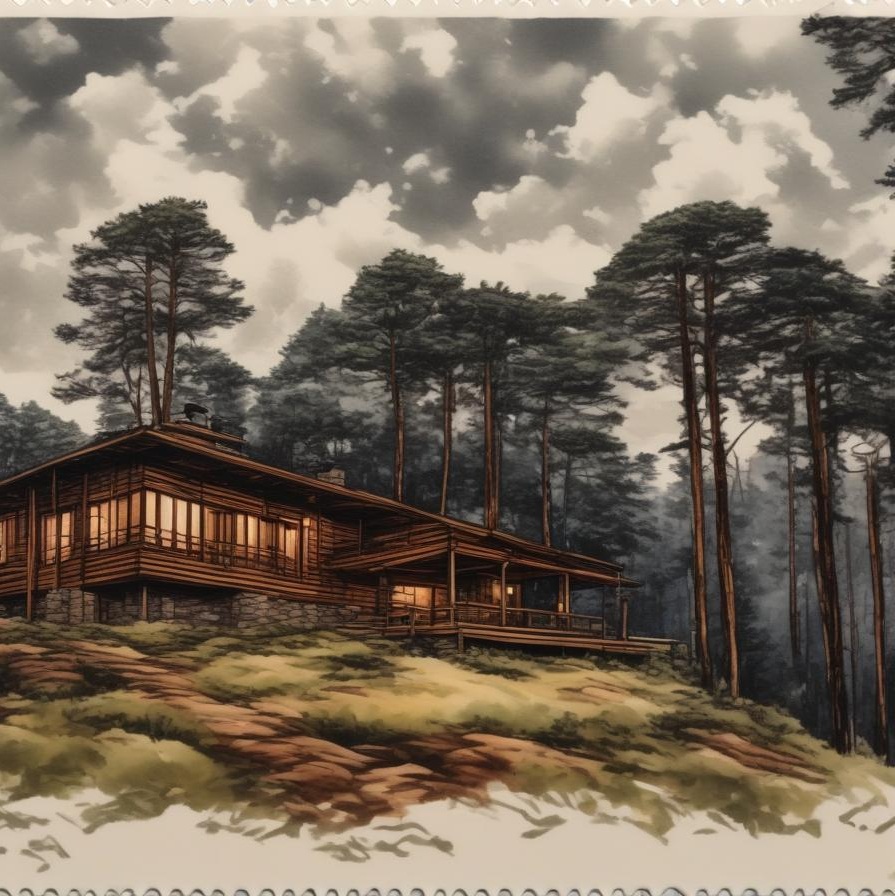} &
                \includegraphics[width=0.16\textwidth]{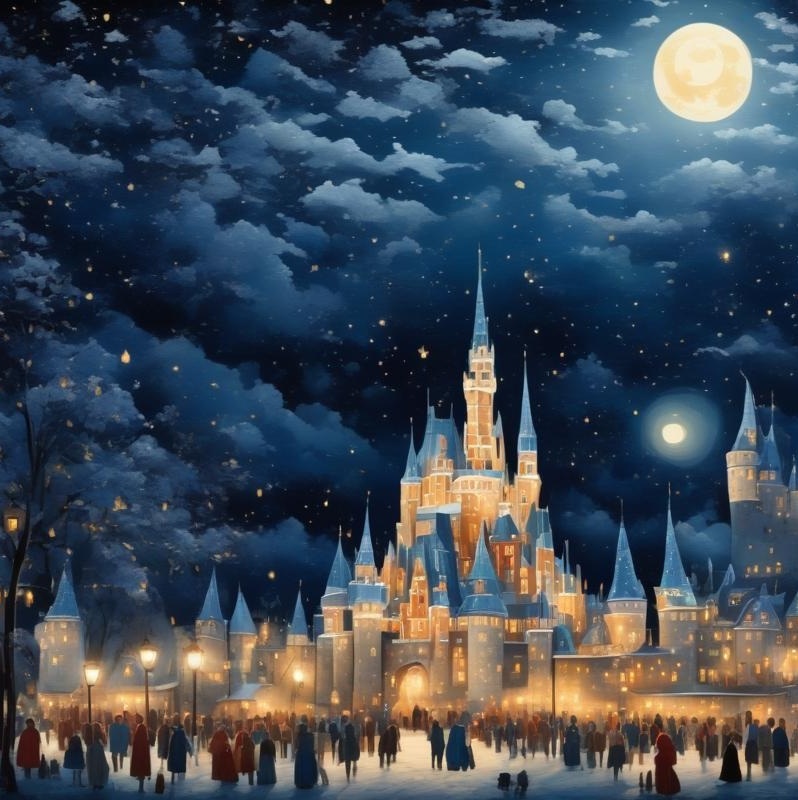} &

        \includegraphics[width=0.32\textwidth]{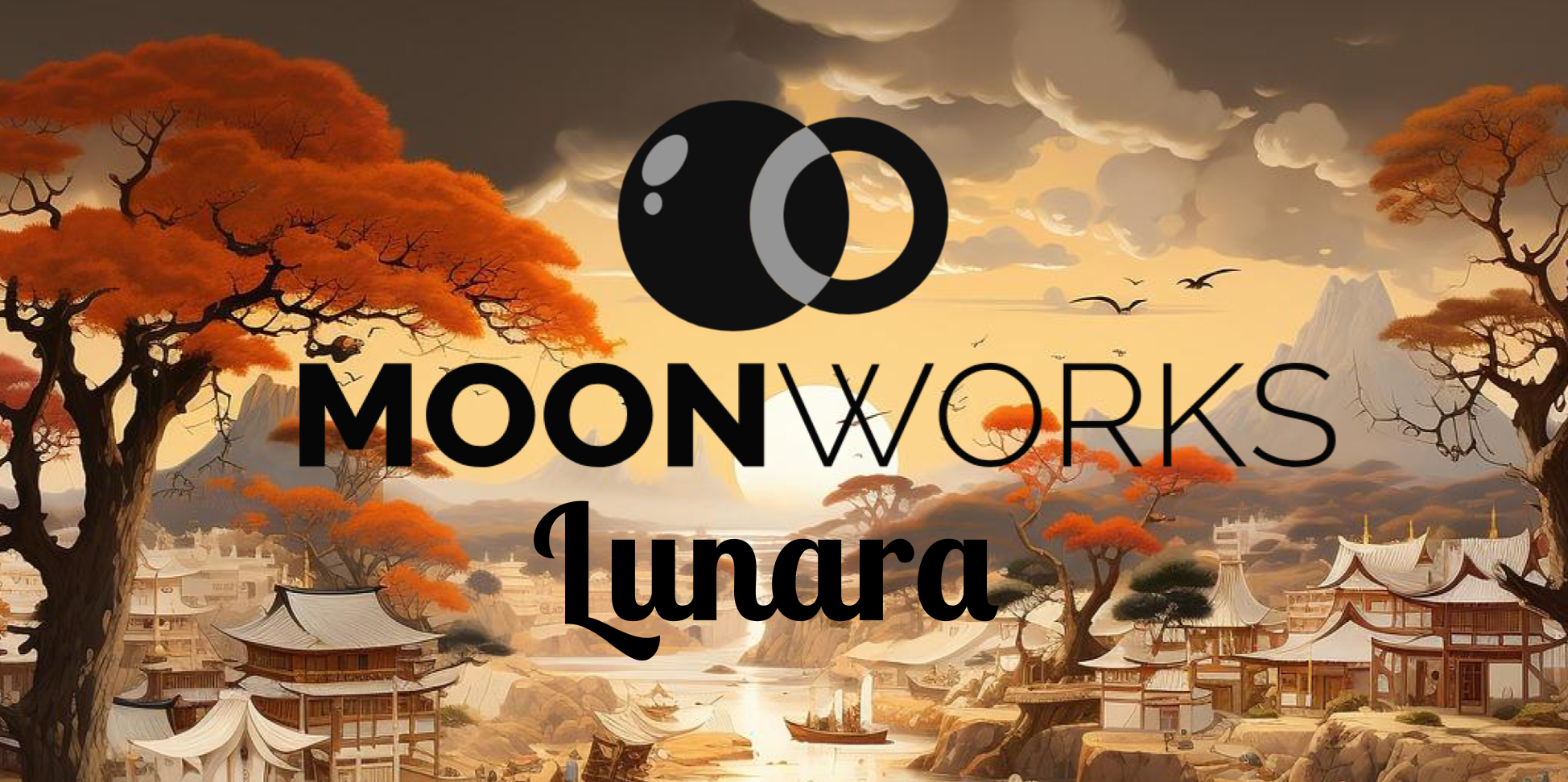}
                \includegraphics[width=0.16\textwidth]{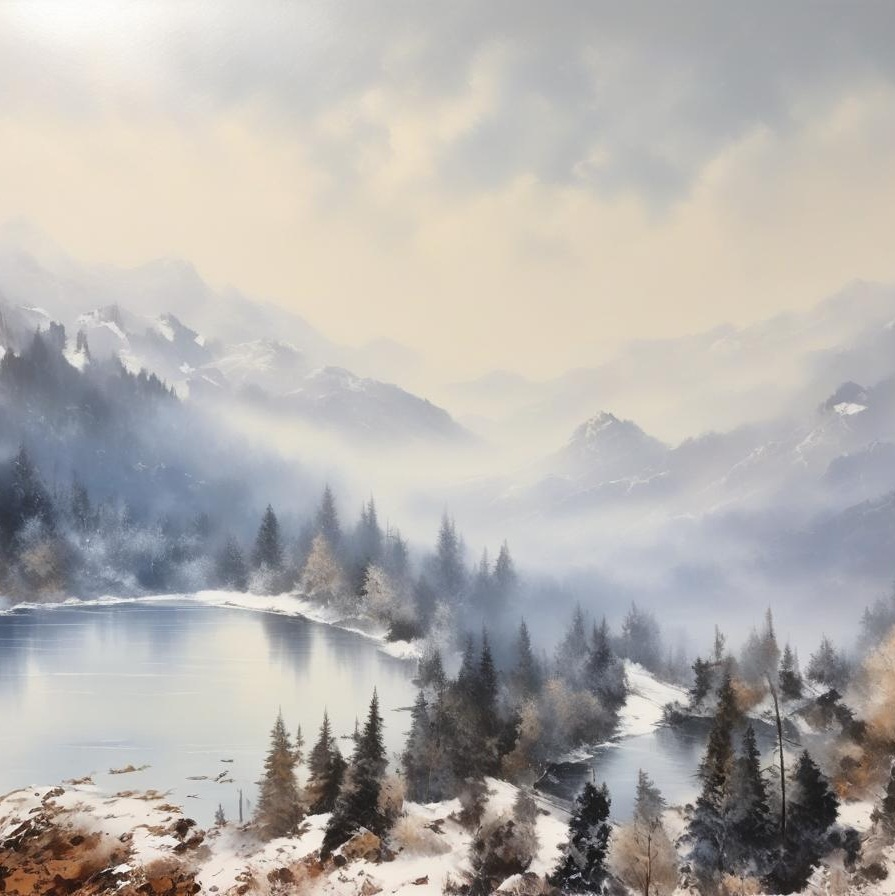} &

        \includegraphics[width=0.16\textwidth]{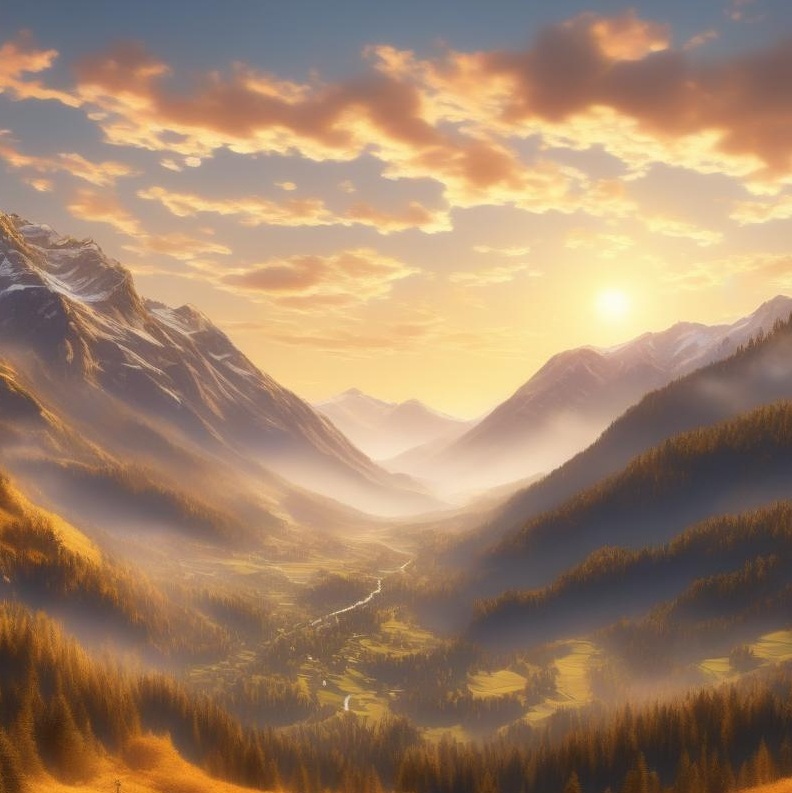}
    \end{tabular}
    \label{fig:style-banner}
\end{figure}

\begin{abstract}

We introduce Lunara Aesthetic II, a publicly released, ethically sourced image dataset designed to support controlled evaluation and learning of contextual consistency in modern image generation and editing systems. The dataset comprises \textbf{2,854} anchor-linked variation pairs derived from \textbf{original} art and photographs created by Moonworks. Each variation pair applies contextual transformations, such as illumination, weather, viewpoint, scene composition, color tone, or mood; while preserving a stable underlying identity. Lunara Aesthetic II operationalizes identity-preserving contextual variation as a supervision signal while also retaining Lunara's signature high aesthetic scores. Results show high identity stability, strong target attribute realization, and a robust aesthetic profile that exceeds large-scale web datasets. Released under the Apache 2.0 license, Lunara Aesthetic II is intended for benchmarking, fine-tuning, and analysis of contextual generalization, identity preservation, and edit robustness in image generation and image-to-image systems with interpretable, relational supervision. The dataset is publicly available at: \url{https://huggingface.co/datasets/moonworks/lunara-aesthetic-image-variations}.

\end{abstract}

\section{Introduction}

Recent frontier text-to-image systems, including OpenAI’s GPT-4o (\cite{openai_4o_image_generation, openai_4o_image_system_card_addendum}) and Google’s Gemini image models (\cite{google_gemini_image_generation_docs}), exhibit substantial advances in instruction following, compositional reasoning, and visual consistency. The outputs of such systems, however, are typically governed by restrictive usage terms that prohibit their use for training or developing competing models. While recent open-source efforts such as Z-image (\cite{team2025zimage}) and Qwen-Image/Edit (\cite{wu2025qwenimagetechnicalreport}) represent important progress, the field continues to lack high-quality, high-aesthetic training data at sufficient signal density. As such, we publicly release Lunara Aesthetic II, a highly curated dataset comprising 2.8K anchor-linked variation sets generated from original artwork, photographs, and the Moonworks Lunara model. The dataset is explicitly designed to operationalize identity-preserving contextual variation, enabling image generation and editing systems to be trained and evaluated on genuine contextual generalization rather than surface-level visual memorization.

Following the release of Lunara Aesthetic I (\cite{wang2026moonworkslunaraaestheticdataset}), which emphasizes aesthetic quality and artistic capabilities, Lunara Aesthetic II extends this work toward fundamental model capabilities such as contextual consistency and semantic transformation, while maintaining high aesthetic quality. 
Lunara Aesthetic II is derived from the original Moonworks image collection and consists entirely of real-world photographs and art contributed with explicit consent for research/commercial use, without personally identifiable information. This supports model development while adhering to ethical data collection and privacy considerations, and demonstrates the viability of ethical, high-signal data construction.

In contrast to large-scale image–text datasets such as LAION-5B (\cite{schuhmann2022laion5b}), which prioritize scale and coverage, and web-harvested caption datasets such as Conceptual Captions and YFCC100M (\cite{sharma2018conceptual, thomee2016yfcc100m}), which lack explicit supervision for identity or contextual consistency, Lunara Aesthetic II is explicitly constructed around anchor-linked, identity-preserving variation. Instruction-based image editing benchmarks, including InstructPix2Pix and MagicBrush (\cite{brooks2023instructpix2pix, zhang2023magicbrush}), introduce text-guided edits and largely consist of isolated, single-step transformations. Lunara Aesthetic II is designed to unify high aesthetic quality with structured, identity-preserving contextual variation, enabling principled training and evaluation of controlled generalization in image generation and editing models.

The contextual variations are generated using Moonworks Lunara, a sub-10B parameter image model employing a novel diffusion mixture architecture trained on similarly structured variation data. Lunara Aesthetic II is organized into explicit classes of visual context change, enabling systematic evaluation of model behavior under controlled transformations. This structure provides substantially stronger learning and evaluation signals than conventional i.i.d. datasets, particularly for assessing identity preservation, attribute controllability, and edit robustness. Both automated metrics and human evaluation validate the quality of the dataset, yielding high identity stability (4.65/5), accurate attribute realization (87.2\% on average), and strong aesthetic performance (mean score 5.9). Consistent with prior Lunara release (\cite{wang2026moonworkslunaraaestheticdataset}), Lunara Aesthetic II is publicly released under the Apache 2.0 license to facilitate reproducible research and broad commercial adoption.



\section{Dataset Overview}

\begin{table}[t]
\centering
\small
\begin{tabular}{l r}
\toprule
\textbf{Statistic} & \textbf{Value} \\
\midrule
Total image pairs (rows) & 2,854 \\
Unique originals & 336 \\
\# topics/ \# of contextual changes & 6/6\\
Variants per original (mean $\pm$ std) & 8.49 $\pm$ 3.26 \\
Variants per original (median / max) & 9 / 15 \\
Original image resolution (median) & 3264$\times$2448 \\
Variant image resolution (median) & 896$\times$1184 \\
Original prompt length (words, mean $\pm$ std) & 14.1 $\pm$ 1.5 \\
Variant prompt length (words, mean $\pm$ std) & 14.6 $\pm$ 2.4 \\
Non-identity variants (rows) & 2,518 \\
contextual changes per variant (mean $\pm$ std) & 2.18 $\pm$ 1.08 \\
contextual changes per variant (median / max) & 2 / 6 \\
\midrule
\midrule
\textbf{contextual change label} & \textbf{\% of rows} \\
\midrule
Illumination-time & 49.68\% \\
Weather-atmosphere & 33.85\% \\
Scene-composition & 33.92\% \\
Mood-atmosphere & 24.21\% \\
Viewpoint-camera & 23.97\% \\
Color-tone & 26.31\% \\
\bottomrule
\end{tabular}
\caption{Summary statistics of the image-variation dataset and frequency of contextual change labels.}
\label{tab:dataset_summary}
\end{table}

Lunara Aesthetic II comprises 2,854 images organized into anchor-linked contextual variation sets derived from 336 original source photographs and arts, explicitly consented for Apache 2.0 release. The dataset is annotated with variation prompts, contextual changes, as well as topics.

\subsection{Dataset Statistics}
\begin{figure}[t]
    \centering
    \includegraphics[width=0.75\textwidth]{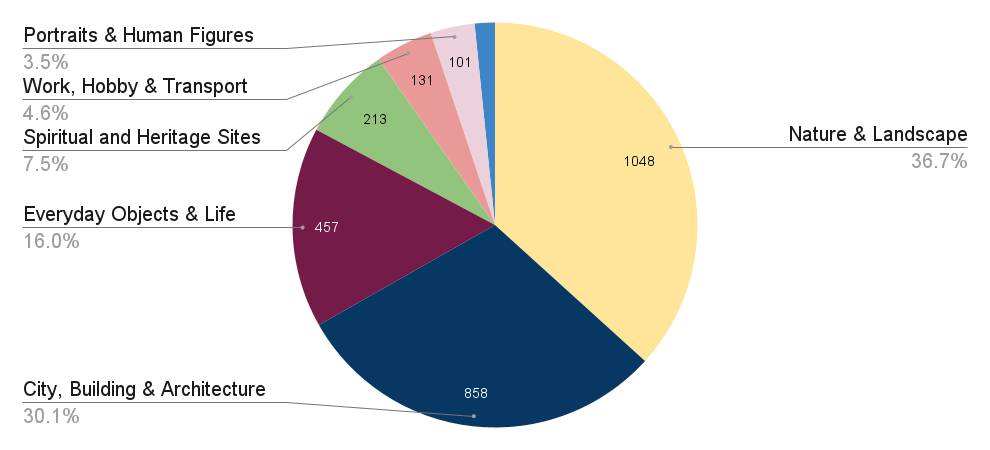}
    \caption{Distribution of topics in the image variation dataset.}
    \label{fig:var-dist}
\end{figure}

\begin{figure}[]
    \centering
    \includegraphics[width=0.65\textwidth]{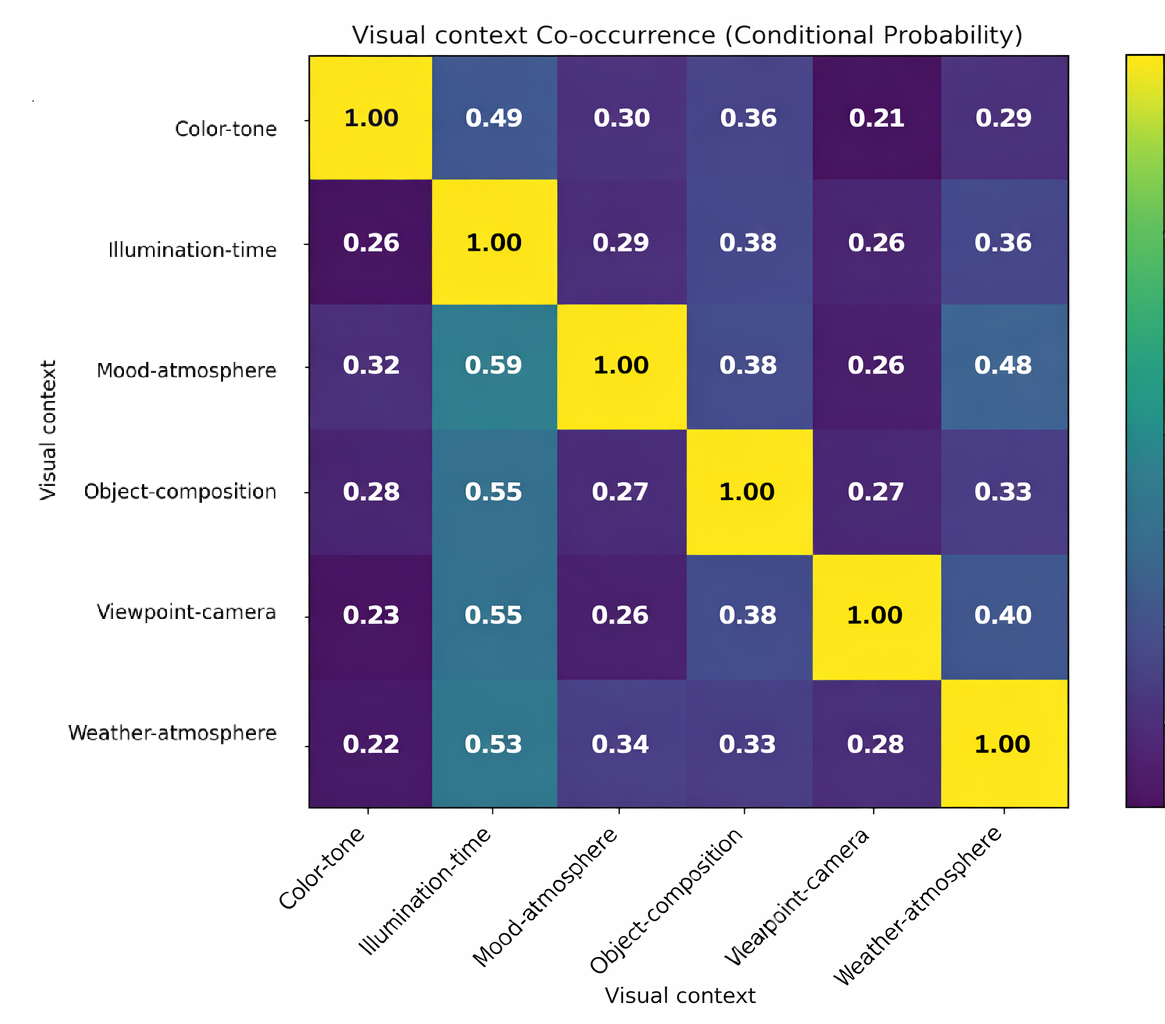}
    \caption{Conditional probability of variations co-occurring.}
    \label{fig:var-con-dist}
\end{figure}
Each contextual variation in the dataset modifies one or more contextual attributes (e.g., illumination, weather, viewpoint, scene composition, color tone, or mood) while preserving the core identity of the source image. Each original is associated with an average of 8.49 variants (std. 3.26), with a median of 9 and a maximum of 15.

Image resolutions vary by source content and variation construction to reflect realistic image editing scenarios rather than enforcing a fixed canonical size. Original images have a median resolution of 3264×2448 (max 3456×4608), while variant (generated) images have a lower median resolution of 896×1184 for practical purposes. 

Prompt complexity remains consistent between originals and variants, with comparable average prompt lengths (14.1 vs. 14.6 words), ensuring that semantic variation is not driven by disproportionate changes in prompt verbosity. Of the 2,854 image pairs, 2,518 correspond to non-identity pairs, while the remainder serve as original source images.

Each variant introduces multiple contextual changes on average (2.18 ± 1.08), with a median of 2 and a maximum of 6, reflecting the compositional nature of the dataset and enabling evaluation of both single-factor and controlled multi-factor contextual interventions.

The dataset spans six topic categories: Nature and Landscape comprising 36.7\% (1,048 images), City, Building, and Architecture 30.1\% (858), Everyday Objects and Daily Life 16.0\% (457), Spiritual and Heritage Sites 7.5\% (213), Work, Hobby, and Transport 4.6\% (131), and Portraits and Human Figures 3.5\% (101).

\subsection{Contextual Change Labels}

The dataset annotations cover a diverse set of contextual change labels (Table~\ref{tab:dataset_summary} and Figure~\ref{fig:var-dist}). Illumination–time changes are the most frequent, accounting for 49.68\% of all rows, followed by scene–composition changes at 33.92\%, reflecting diversity in objects and living subjects within the scene. Weather–atmosphere changes appear in 33.85\% of instances, followed by color–tone changes at 26.31\%. 
More subjective or contextual changes, such as mood–atmosphere and viewpoint–camera, occur in 24.21\% and 23.97\% of rows, respectively. Note that rows may be associated with multiple contextual change labels, and percentages therefore do not sum to 100\%.

\subsection{Conditional Co-occurrence of contextual Variations}

We analyze the conditional co-occurrence structure of contextual variation labels in the dataset (see Figure~\ref{fig:var-con-dist}). Substantial off-diagonal mass indicates non-trivial dependencies between change types. Illumination–time emerges as a central factor, exhibiting consistently high conditional probabilities with mood–atmosphere, object–composition, viewpoint–camera, and weather–atmosphere changes (approximately 0.53–0.59), suggesting that lighting variations frequently accompany other contextual edits. Scene-level and atmospheric categories show moderate coupling with one another (roughly 0.33–0.48), reflecting coordinated broad-level modifications. Color–tone displays selective coupling—most notably with illumination–time (~0.49), while remaining more weakly associated with other categories. Viewpoint–camera changes exhibit moderate associations with object–composition and weather–atmosphere, indicating that geometric adjustments often co-occur with broader scene alterations. Overall, these patterns reveal structured, non-uniform dependencies among contextual variations, highlighting the compositional nature of the dataset.




\begin{figure}[t]
    \centering
    \includegraphics[width=0.9\textwidth]{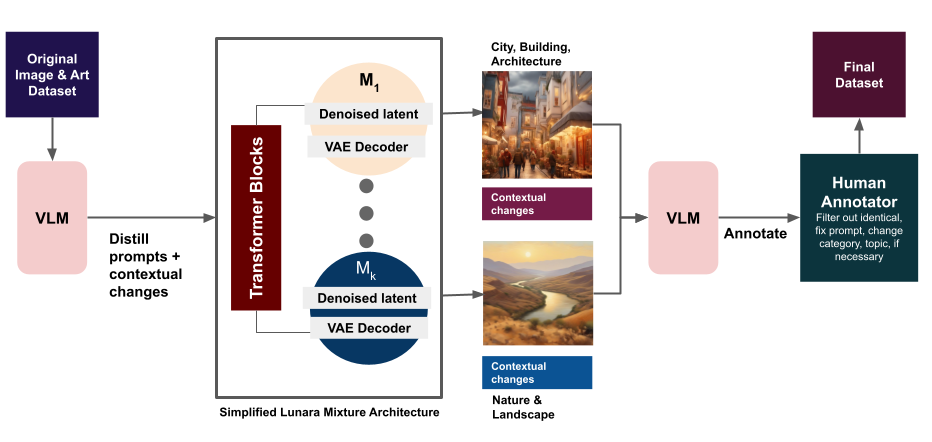}
    \caption{Distribution of variation types in the image variation dataset.}
    \label{fig:var-dist}
\end{figure}

\begin{figure}[]
    \centering

    \begin{subfigure}[t]{0.48\textwidth}
        \centering
        \begin{minipage}{0.48\textwidth}
            \centering
            \includegraphics[height=2.8cm, keepaspectratio]{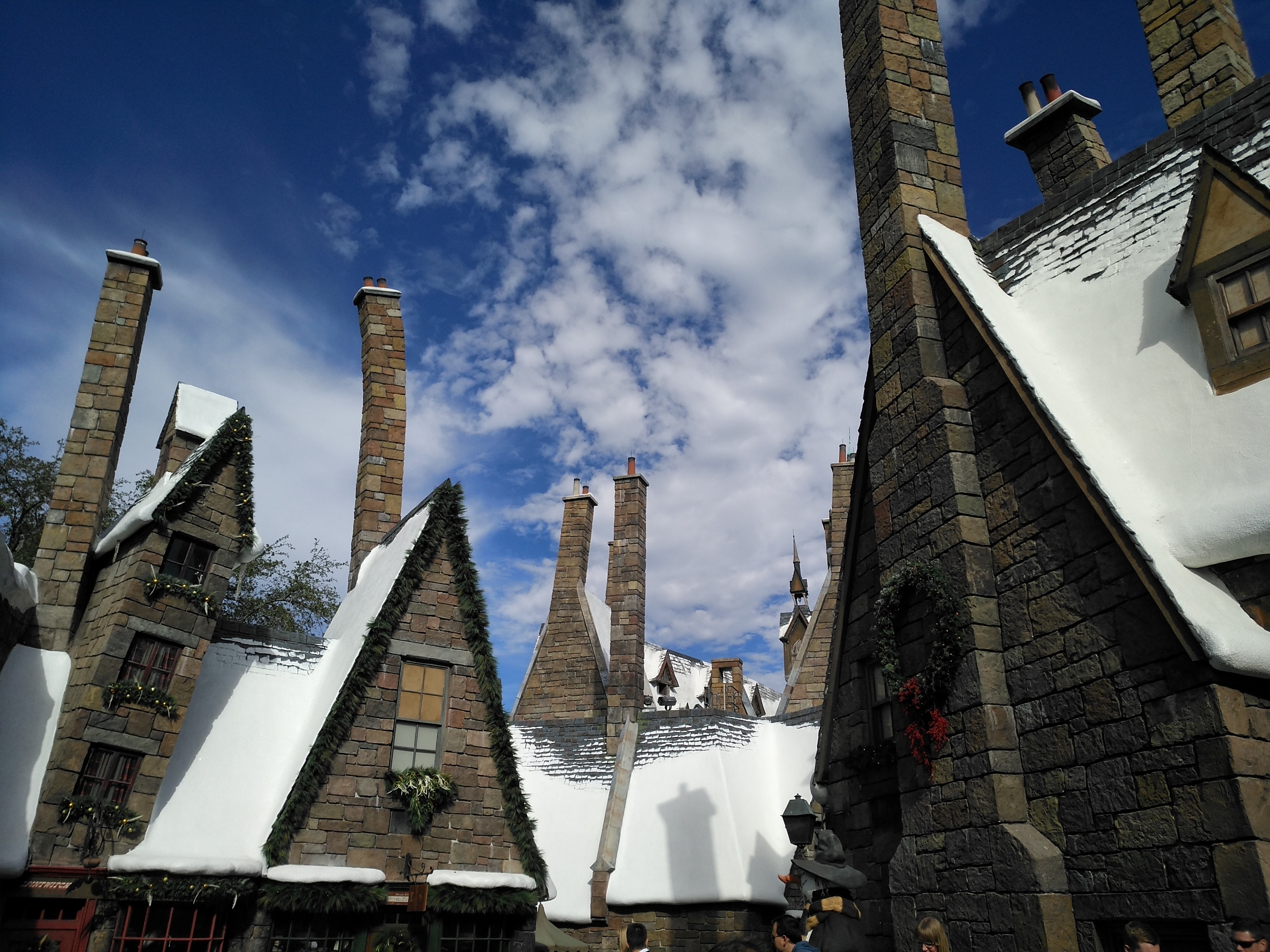}
            \vspace{-0.4em}

            {\scriptsize\raggedright Hogsmede houses with snow.}
        \end{minipage}
        \hfill
        \begin{minipage}{0.48\textwidth}
            \centering
            \includegraphics[height=2.8cm, keepaspectratio]{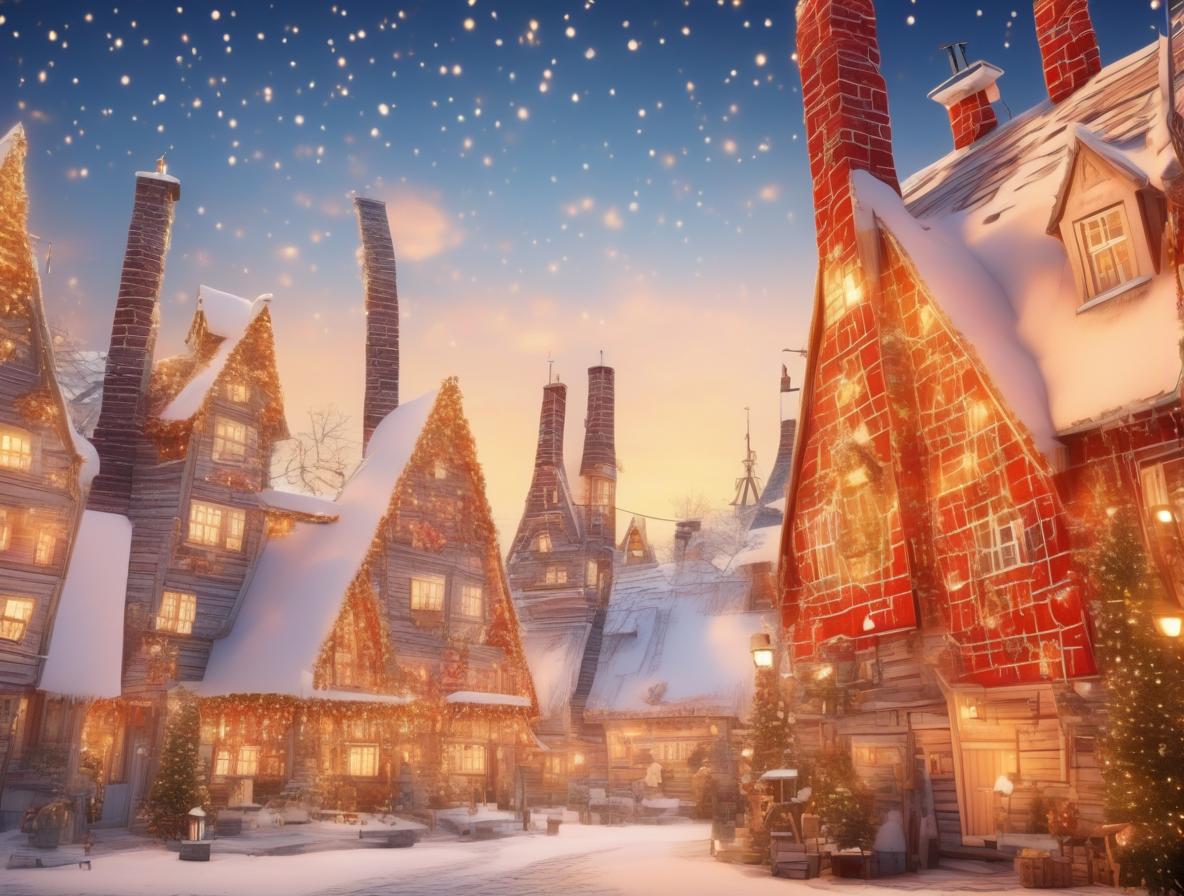}
            \vspace{-0.4em}

            {\scriptsize\raggedright Hogsmede in Christmas.}
        \end{minipage}
        \caption{Change of illumination time and mood}
    \end{subfigure}
    \hfill
    \begin{subfigure}[t]{0.48\textwidth}
        \centering
        \begin{minipage}{0.48\textwidth}
            \centering
            \includegraphics[height=2.8cm, keepaspectratio]{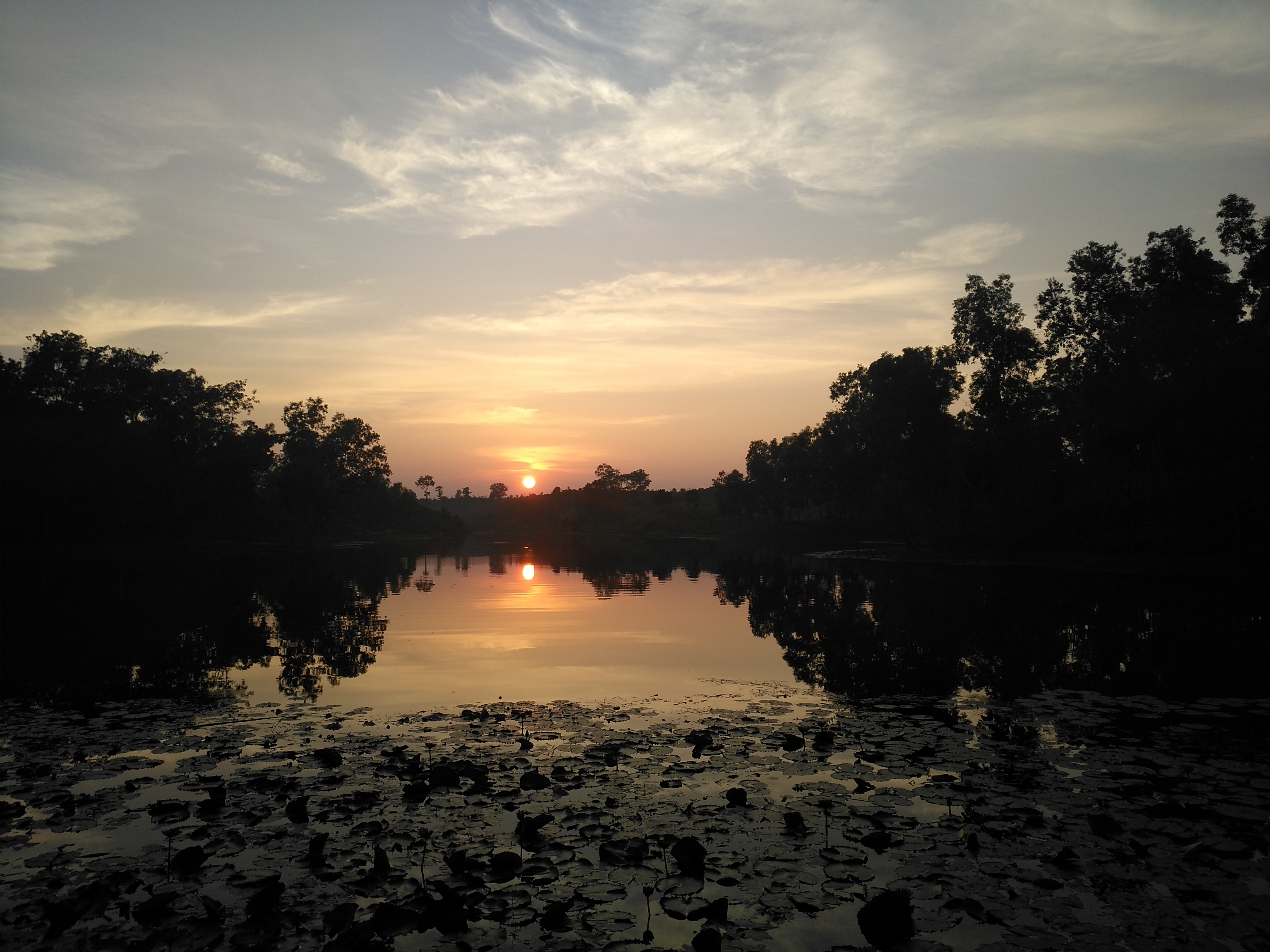}
            \vspace{-0.4em}

            {\scriptsize\raggedright Sunset on lake with lilies.}
        \end{minipage}
        \hfill
        \begin{minipage}{0.48\textwidth}
            \centering
            \includegraphics[height=2.8cm, keepaspectratio]{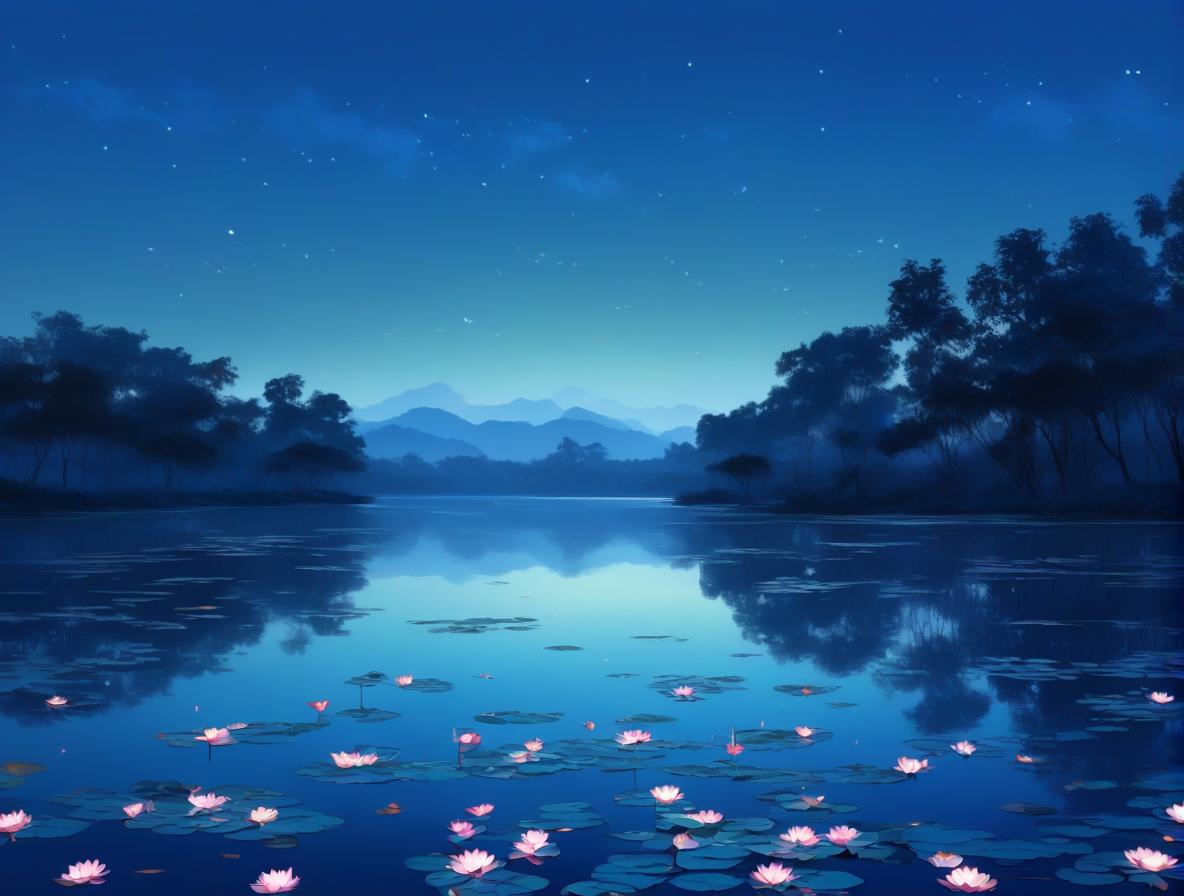}
            \vspace{-0.4em}

            {\scriptsize\raggedright Lake at night with lilies.}
        \end{minipage}
        \caption{Change of illumination time}
    \end{subfigure}

    \vspace{0.6em}

    \begin{subfigure}[t]{0.48\textwidth}
        \centering
        \begin{minipage}{0.48\textwidth}
            \centering
            \includegraphics[height=2.8cm, keepaspectratio]{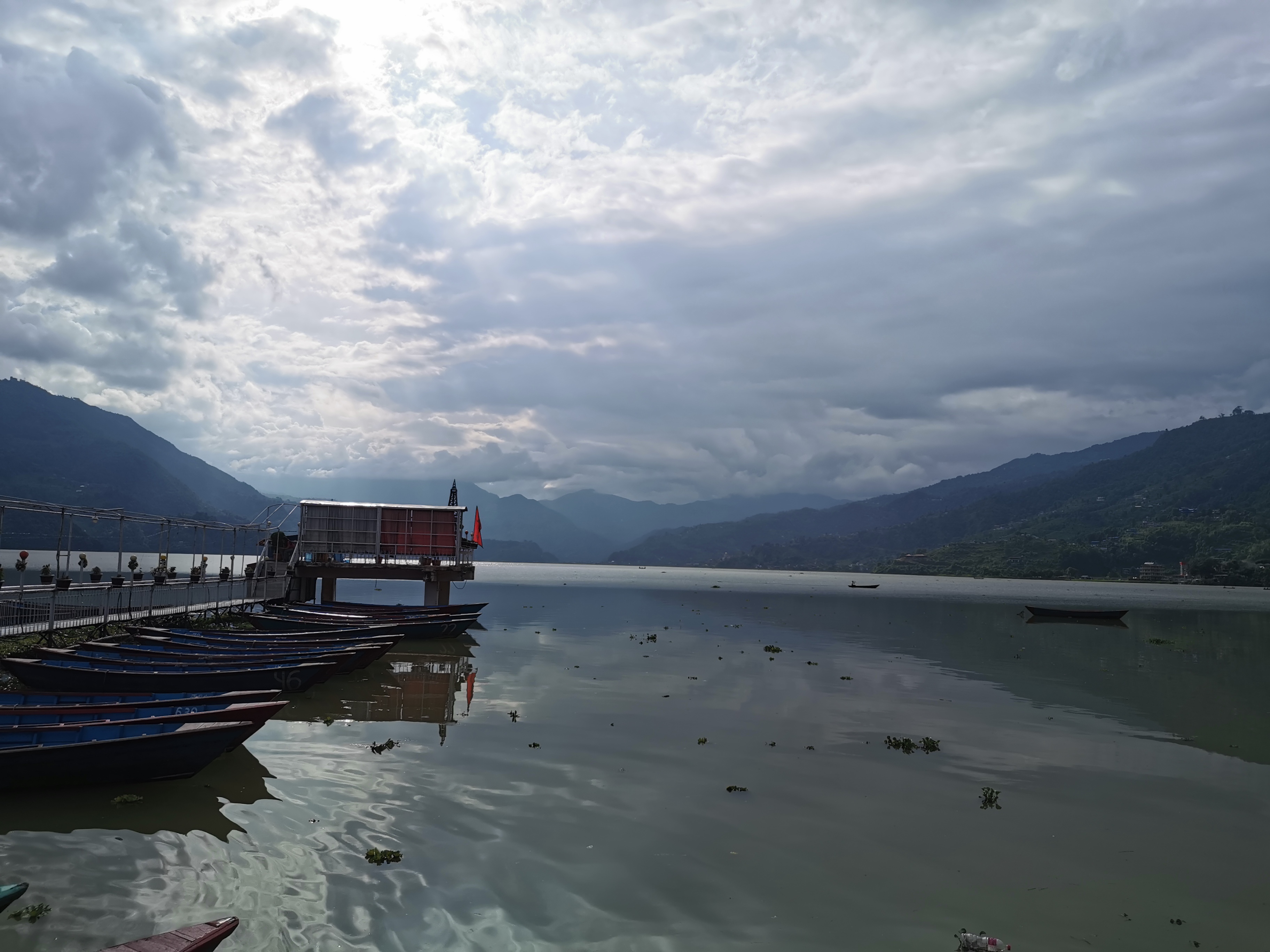}
            \vspace{-0.4em}

            {\scriptsize\raggedright Serene lake with boats..}
        \end{minipage}
        \hfill
        \begin{minipage}{0.48\textwidth}
            \centering
            \includegraphics[height=2.8cm, keepaspectratio]{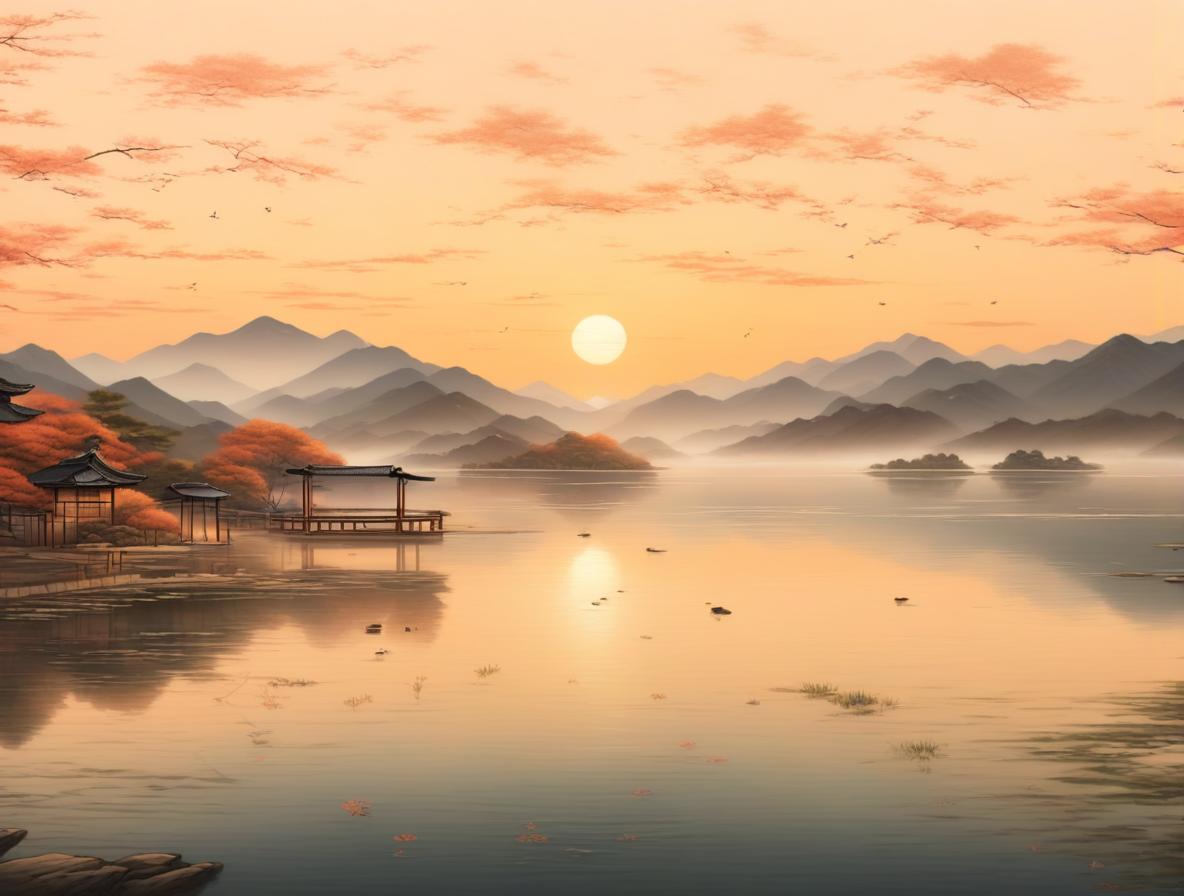}
            \vspace{-0.4em}

            {\scriptsize\raggedright Serene lake in autumn..}
        \end{minipage}
        \caption{Change of weather and scene-composition}
    \end{subfigure}
    \hfill
    \begin{subfigure}[t]{0.48\textwidth}
        \centering
        \begin{minipage}{0.48\textwidth}
            \centering
            \includegraphics[height=2.8cm, keepaspectratio]{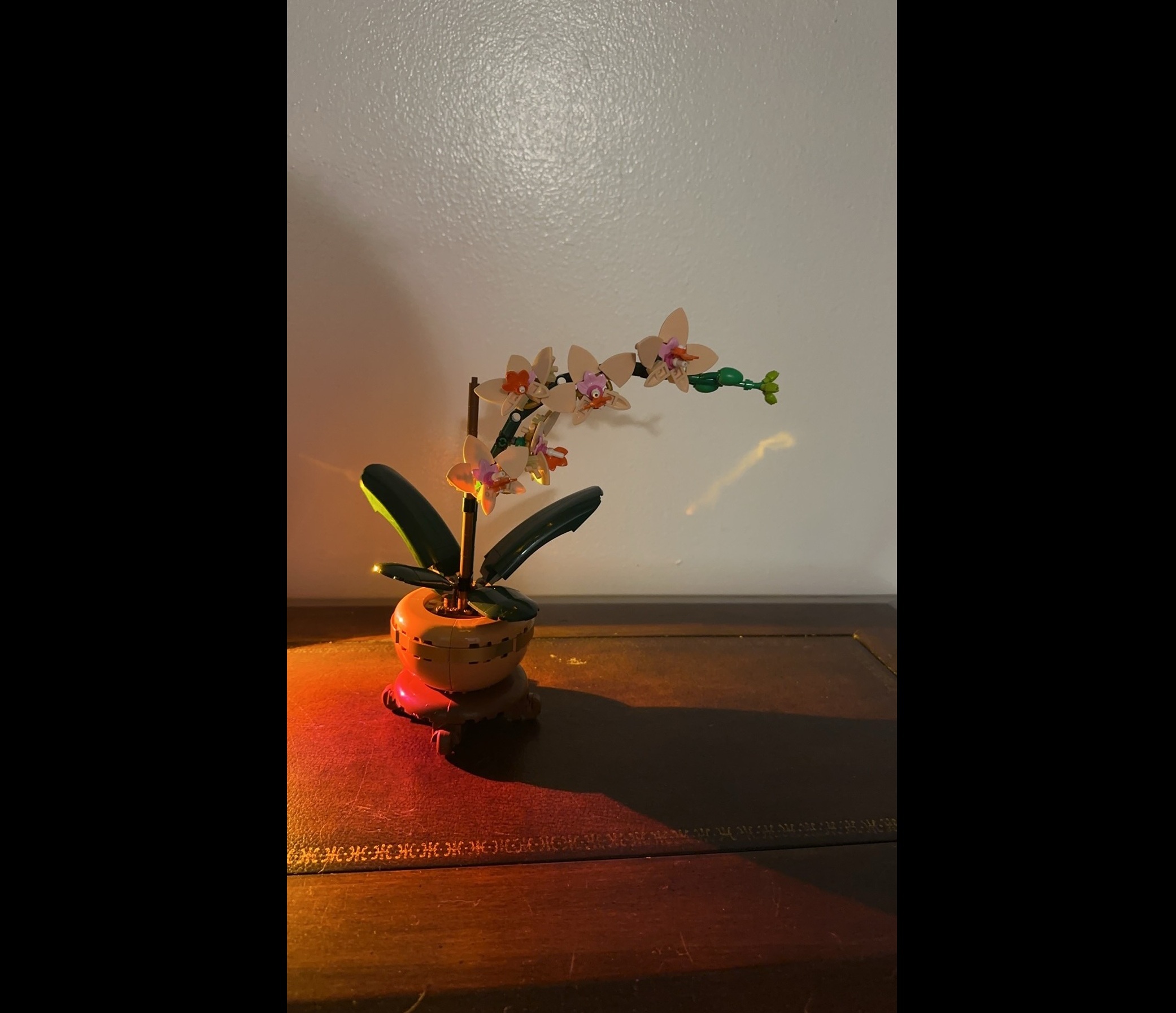}
            \vspace{-0.4em}

            {\scriptsize\raggedright Flower vase on table..}
        \end{minipage}
        \hfill
        \begin{minipage}{0.48\textwidth}
            \centering
            \includegraphics[height=2.8cm, keepaspectratio]{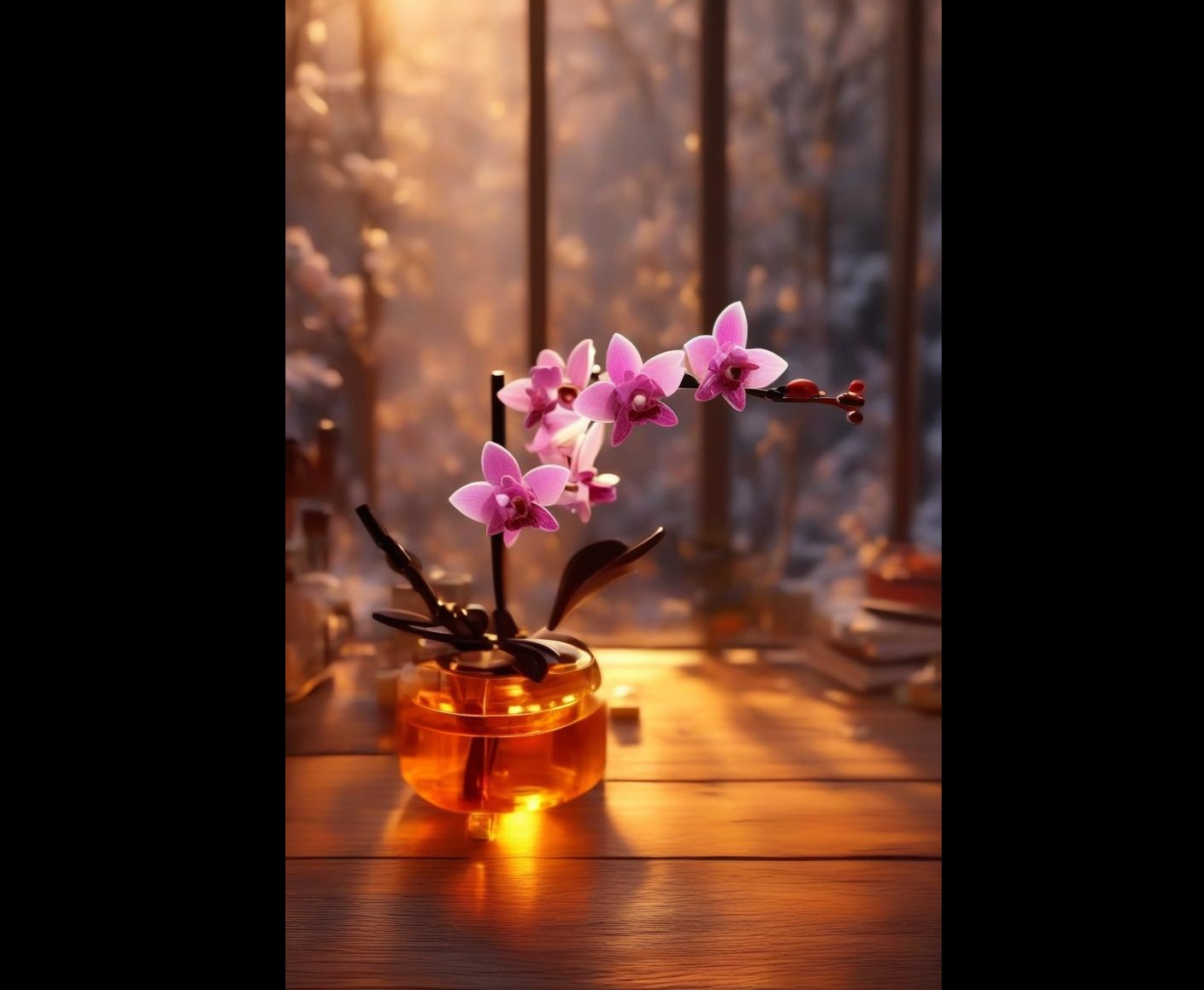}
            \vspace{-0.4em}

            {\scriptsize\raggedright Flower vase near window..}
        \end{minipage}
        \caption{Change of scene-composition and color tone}
    \end{subfigure}

    \vspace{0.6em}

    \begin{subfigure}[t]{0.48\textwidth}
        \centering
        \begin{minipage}{0.48\textwidth}
            \centering
            \includegraphics[height=2.8cm, keepaspectratio]{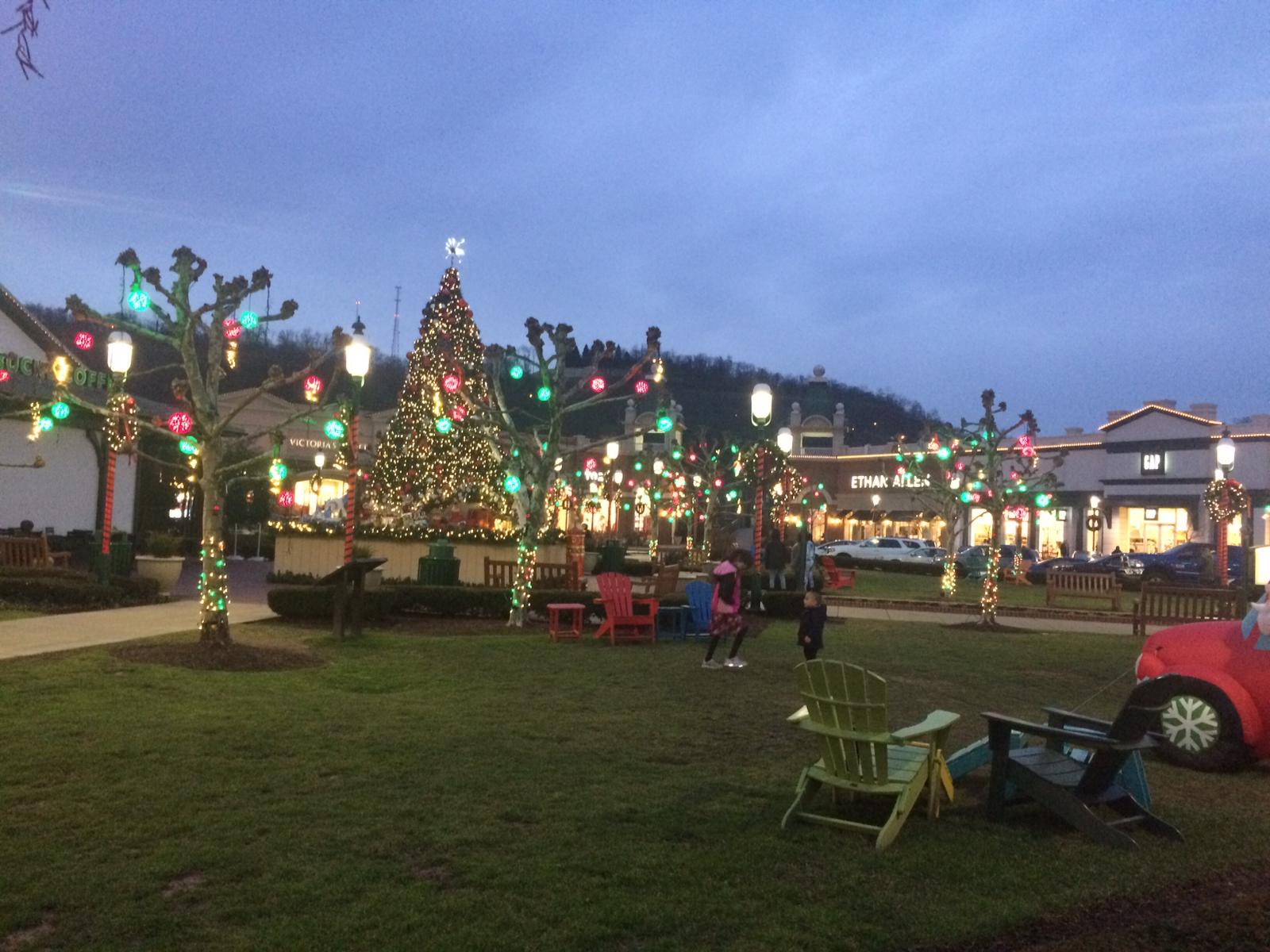}
            \vspace{-0.4em}

            {\scriptsize\raggedright Christmas trees in park..}
        \end{minipage}
        \hfill
        \begin{minipage}{0.48\textwidth}
            \centering
            \includegraphics[height=2.8cm, keepaspectratio]{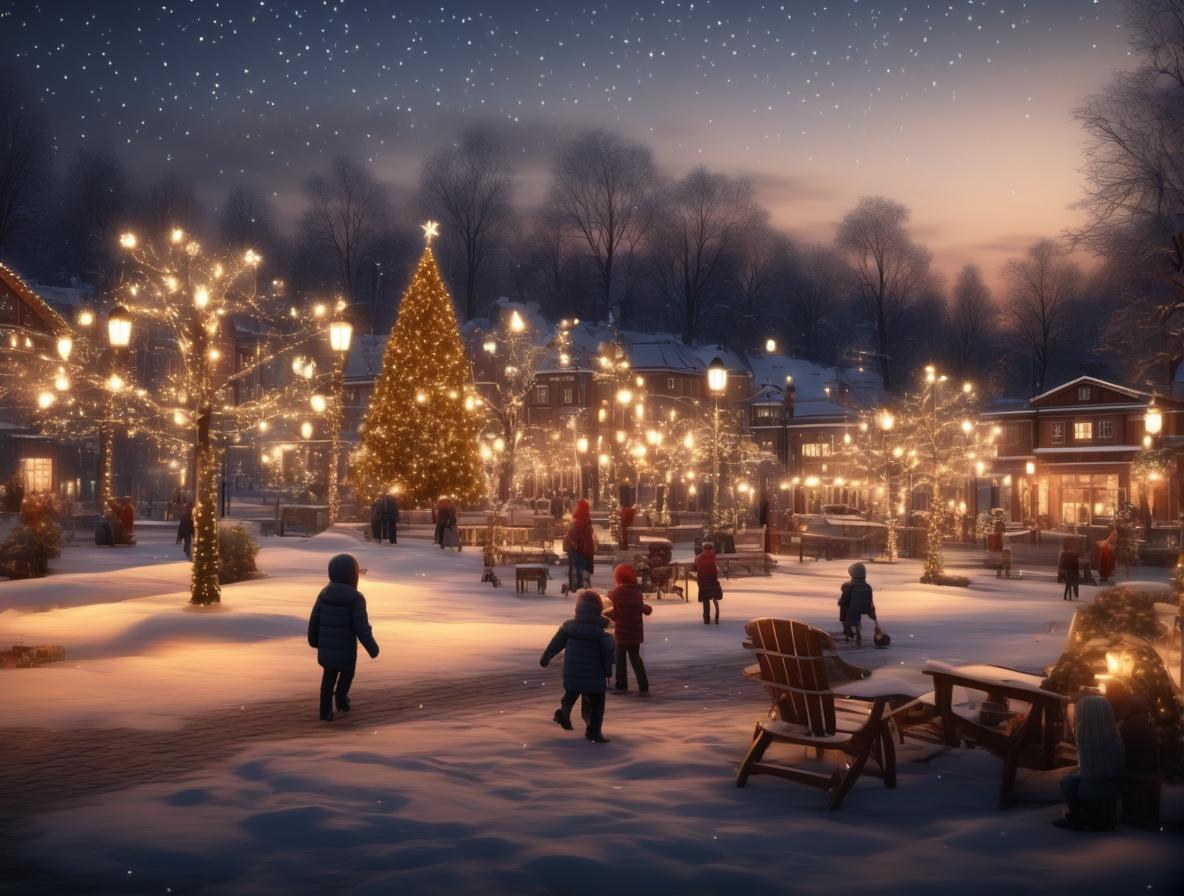}
            \vspace{-0.4em}

            {\scriptsize\raggedright Christmas trees in snow..}
        \end{minipage}
        \caption{Change of viewpoint-camera and weather}
    \end{subfigure}
    \hfill
    \begin{subfigure}[t]{0.48\textwidth}
        \centering
        \begin{minipage}{0.48\textwidth}
            \centering
            \includegraphics[height=2.8cm, keepaspectratio]{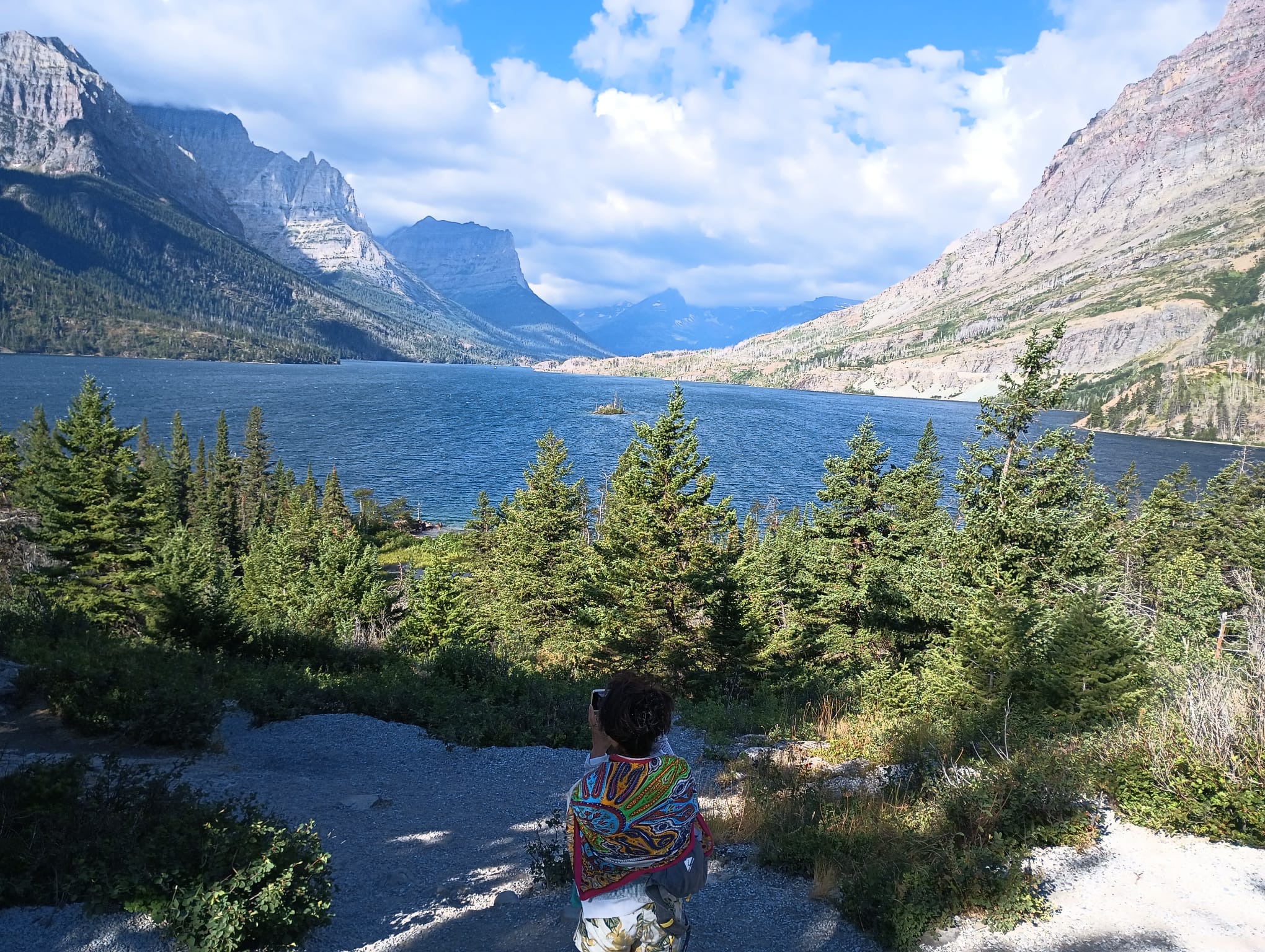}
            \vspace{-0.4em}

            {\scriptsize\raggedright Looking over a lake..}
        \end{minipage}
        \hfill
        \begin{minipage}{0.48\textwidth}
            \centering
            \includegraphics[height=2.8cm, keepaspectratio]{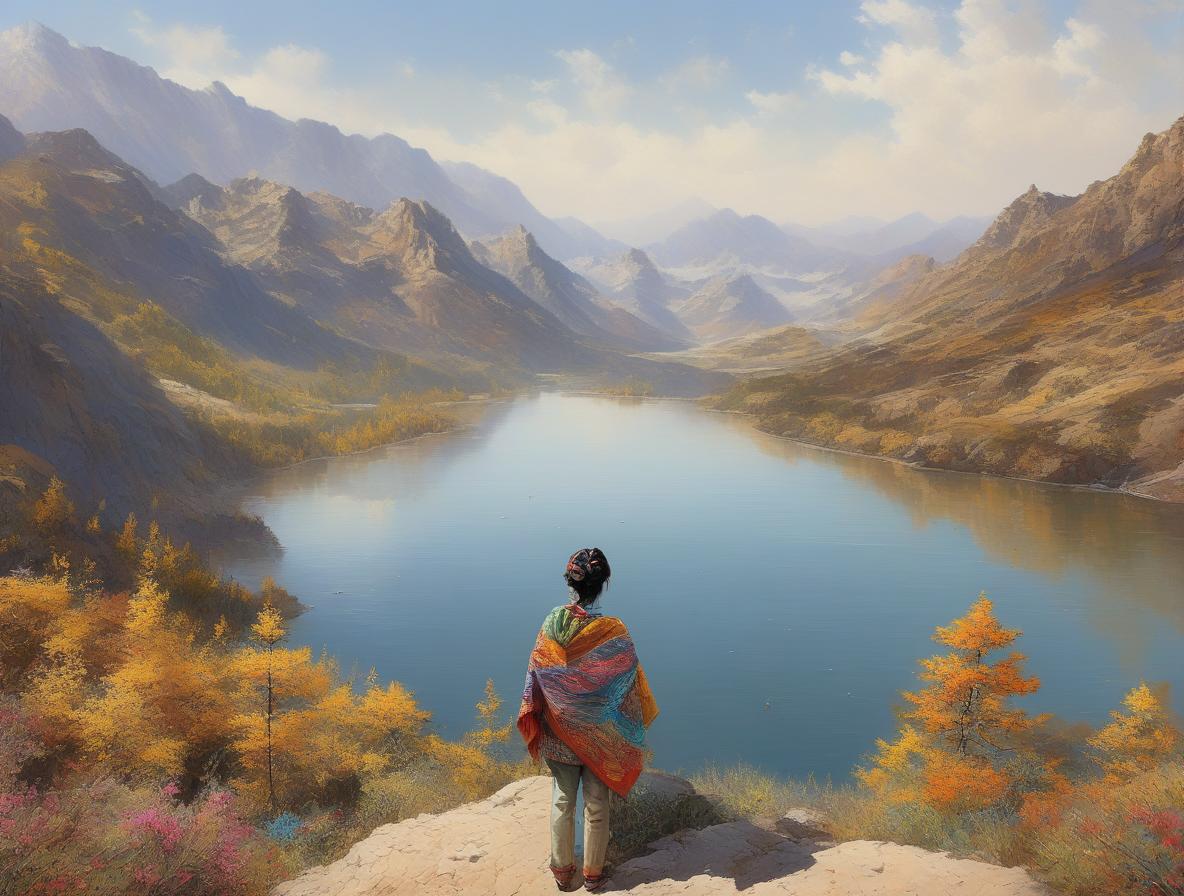}
            \vspace{-0.4em}

            {\scriptsize\raggedright From further away, in fall..}
        \end{minipage}
        \caption{Change of weather}
    \end{subfigure}

    \caption{Contextual variations in \textbf{original photographs}. Each pair shows original (left) and modified image (right).}
    \label{fig:contextual-variations-photo}
\end{figure}


\begin{figure}[]
    \centering

    \begin{subfigure}[t]{0.48\textwidth}
        \centering
        \begin{minipage}[t]{0.48\textwidth}
            \vspace{0pt}
            \centering
            \includegraphics[height=2.8cm, keepaspectratio]{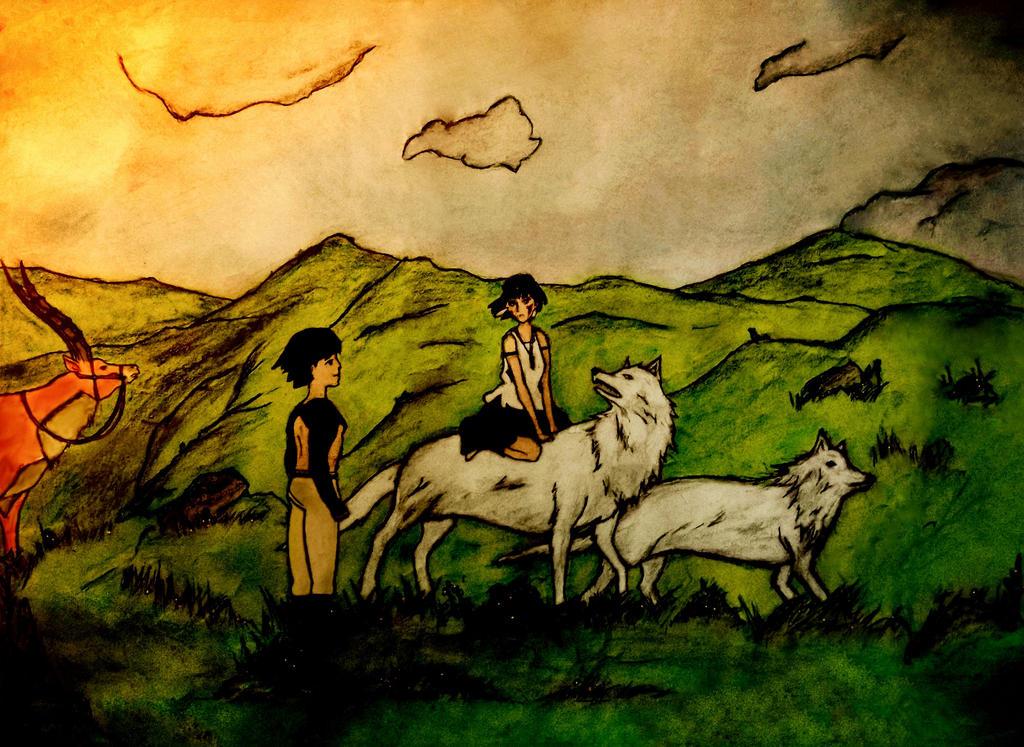}
            \vspace{-0.4em}

            {\scriptsize\raggedright two figures and wolves..}
        \end{minipage}
        \hfill
        \begin{minipage}[t]{0.48\textwidth}
            \vspace{0pt}
            \centering
            \includegraphics[height=2.8cm, keepaspectratio]{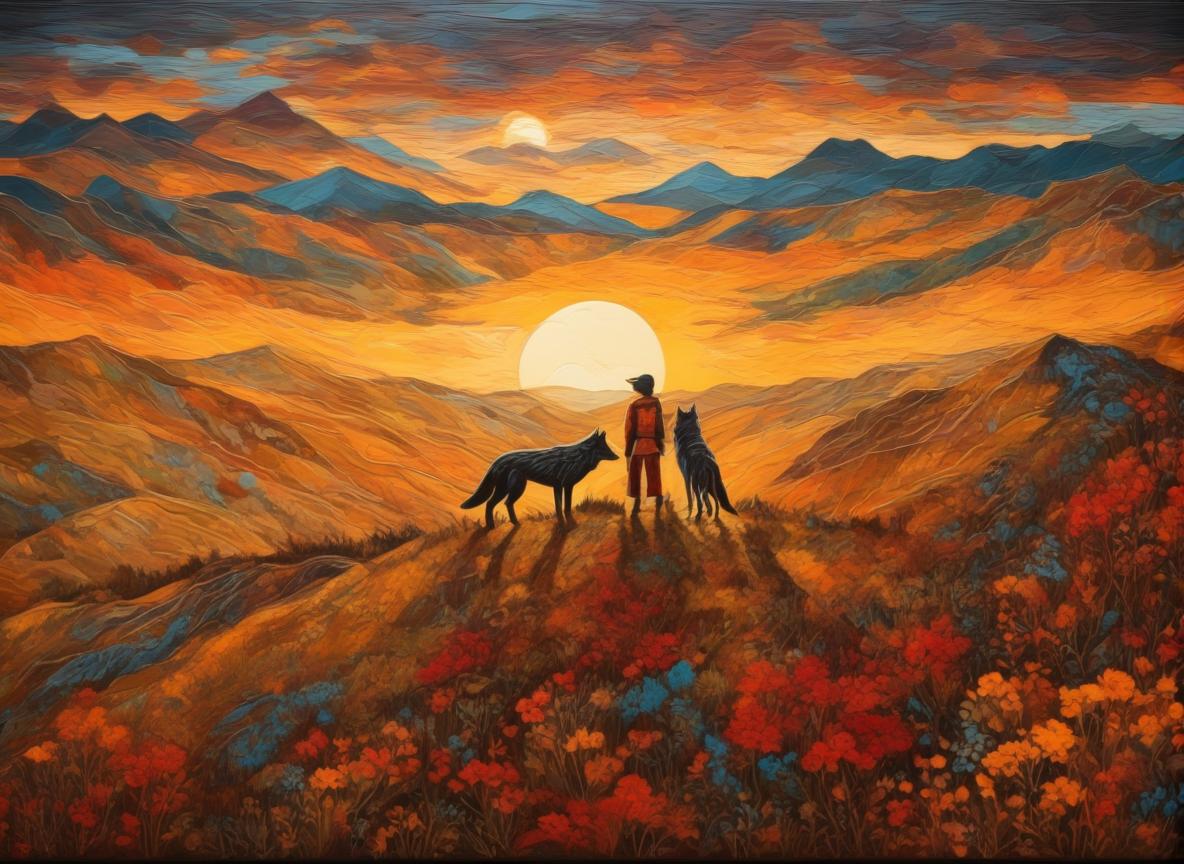}
            \vspace{-0.4em}

            {\scriptsize\raggedright a figure and two wolves..}
        \end{minipage}
        \caption{Change of scene-composition}
    \end{subfigure}
    \hfill
    \begin{subfigure}[t]{0.48\textwidth}
        \centering
        \begin{minipage}[t]{0.48\textwidth}
            \vspace{0pt}
            \centering
            \includegraphics[height=2.8cm, keepaspectratio]{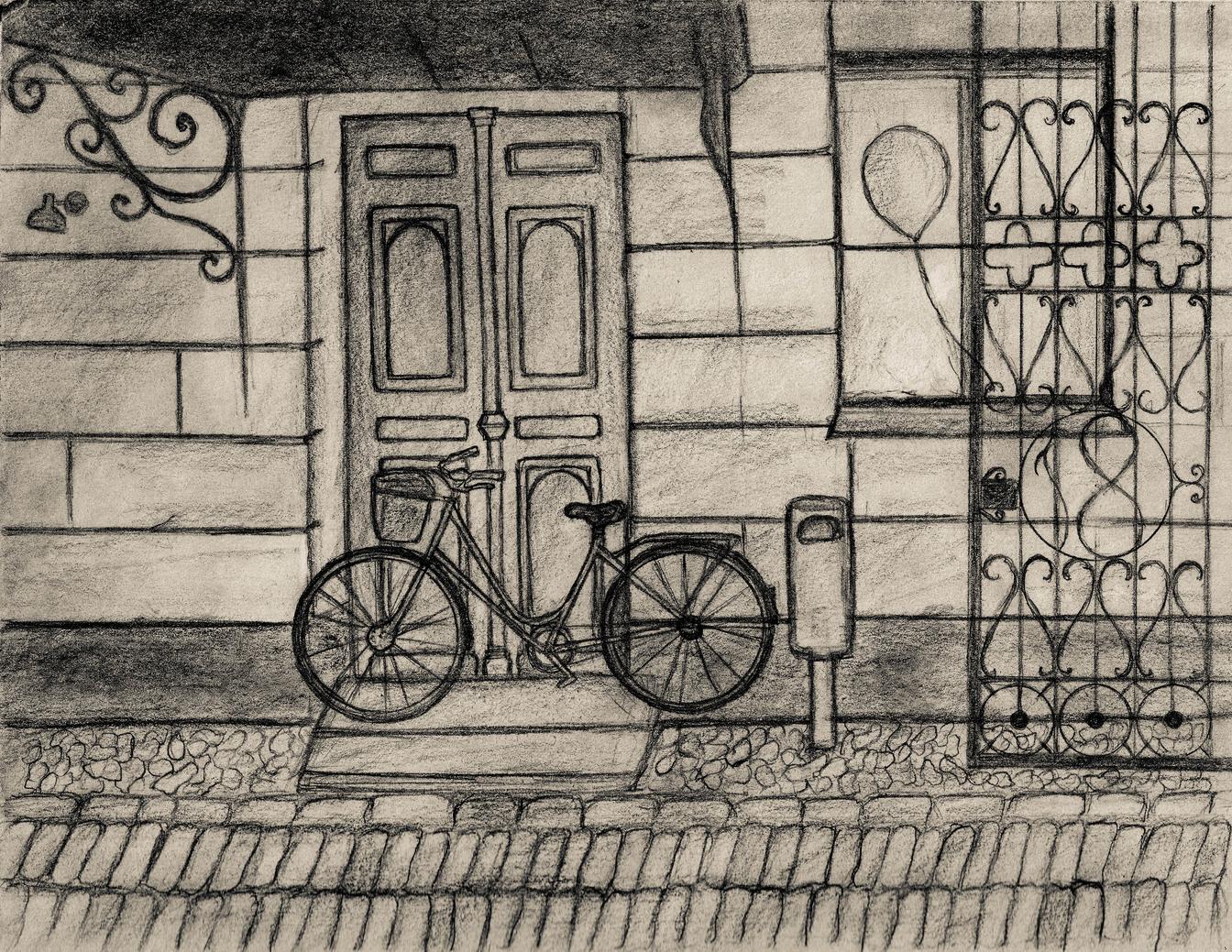}
            \vspace{-0.4em}

            {\scriptsize\raggedright bicycle in front of..}
        \end{minipage}
        \hfill
        \begin{minipage}[t]{0.48\textwidth}
            \vspace{0pt}
            \centering
            \includegraphics[height=2.8cm, keepaspectratio]{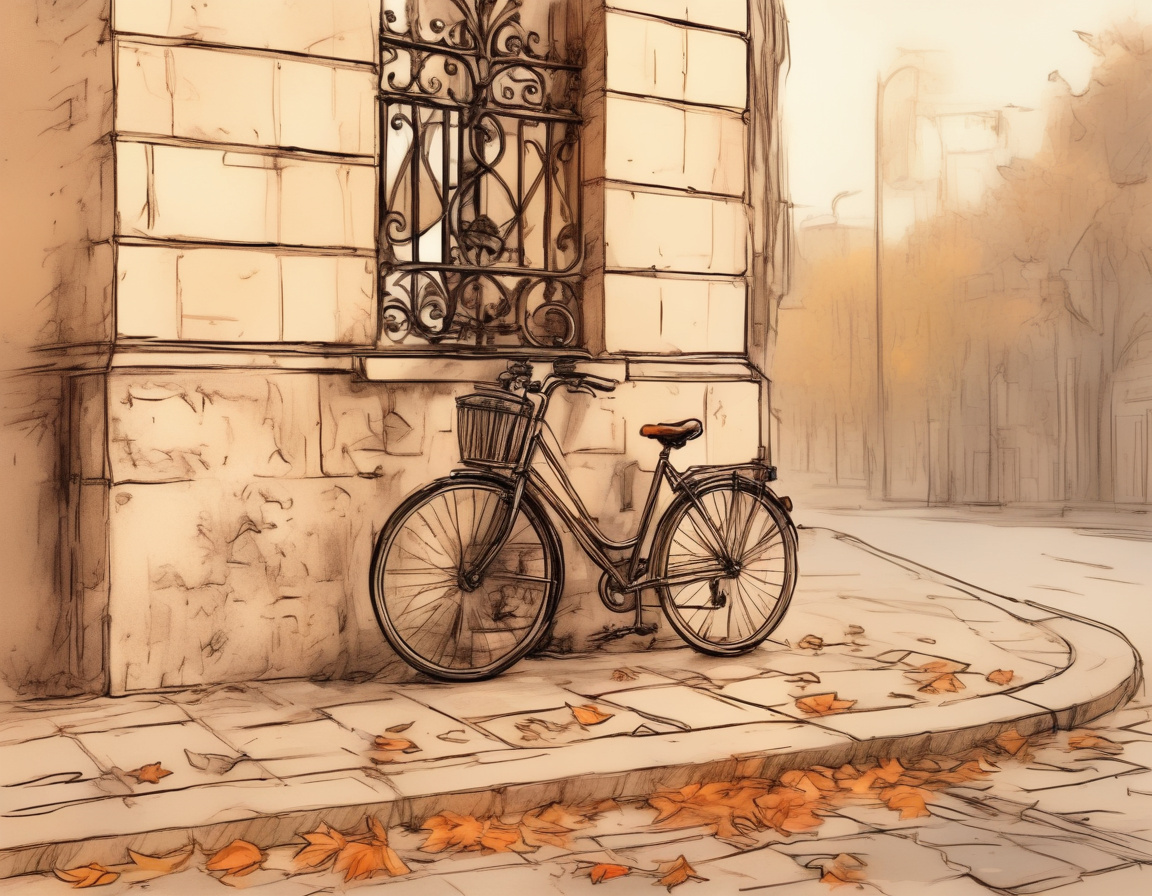}
            \vspace{-0.4em}

            {\scriptsize\raggedright bicycle around corner..}
        \end{minipage}
        \caption{Change of viewpoint camera + color-tone}
    \end{subfigure}

    \vspace{0.6em}

    \begin{subfigure}[t]{0.48\textwidth}
        \centering
        \begin{minipage}[t]{0.48\textwidth}
            \vspace{0pt}
            \centering
            \includegraphics[height=2.8cm, keepaspectratio]{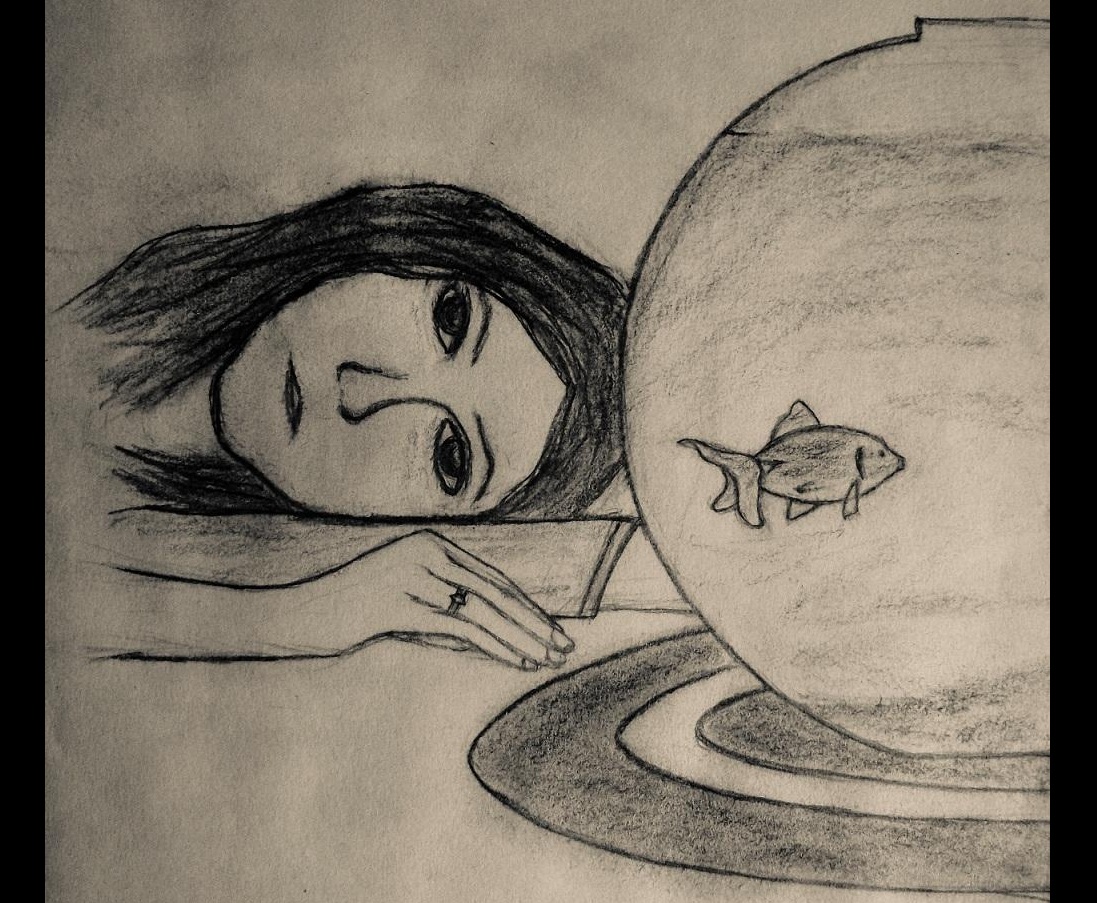}
            \vspace{0.74em}

            {\scriptsize\raggedright next to a goldfish bowl..}
        \end{minipage}
        \hfill
        \begin{minipage}[t]{0.48\textwidth}
            \vspace{0pt}
            \centering
            \includegraphics[height=2.8cm, keepaspectratio]{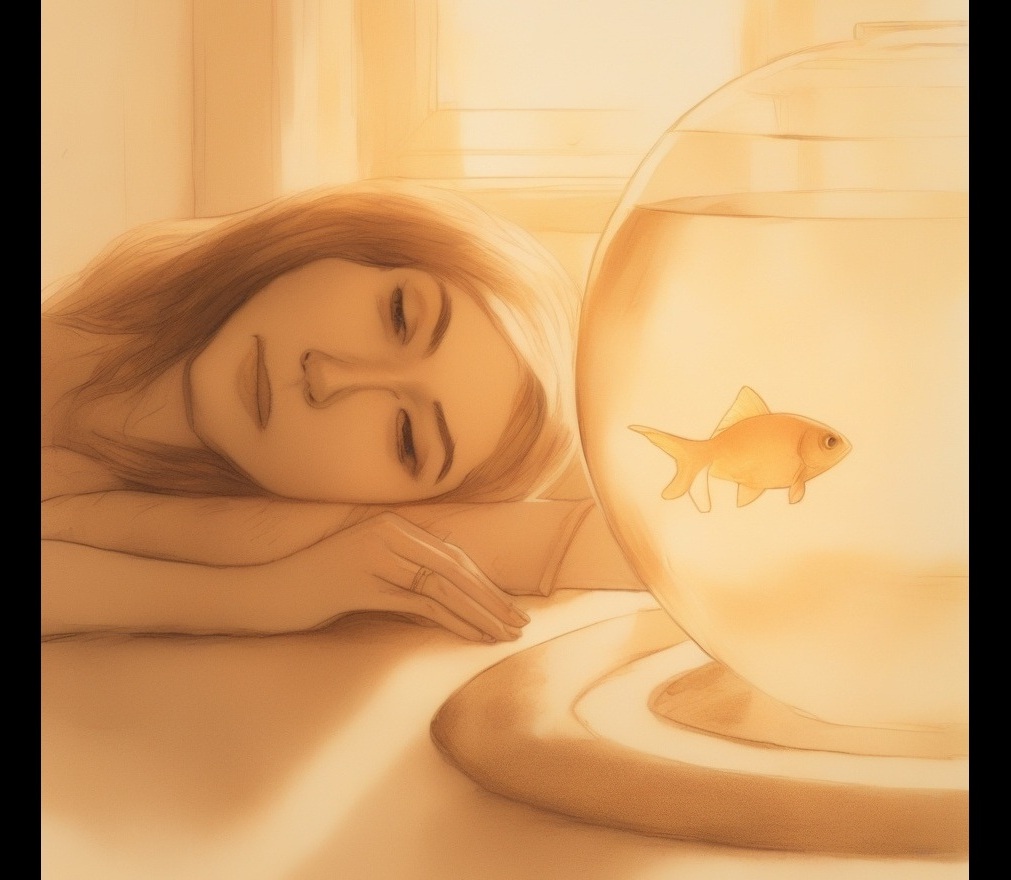}
            \vspace{0.74em}

            {\scriptsize\raggedright warm light and mood..}
        \end{minipage}
        \caption{Change of illumination time and mood}
    \end{subfigure}
    \hfill
    \begin{subfigure}[t]{0.48\textwidth}
        \centering
        \begin{minipage}[t]{0.48\textwidth}
            \vspace{0pt}
            \centering
            \includegraphics[height=2.8cm, keepaspectratio]{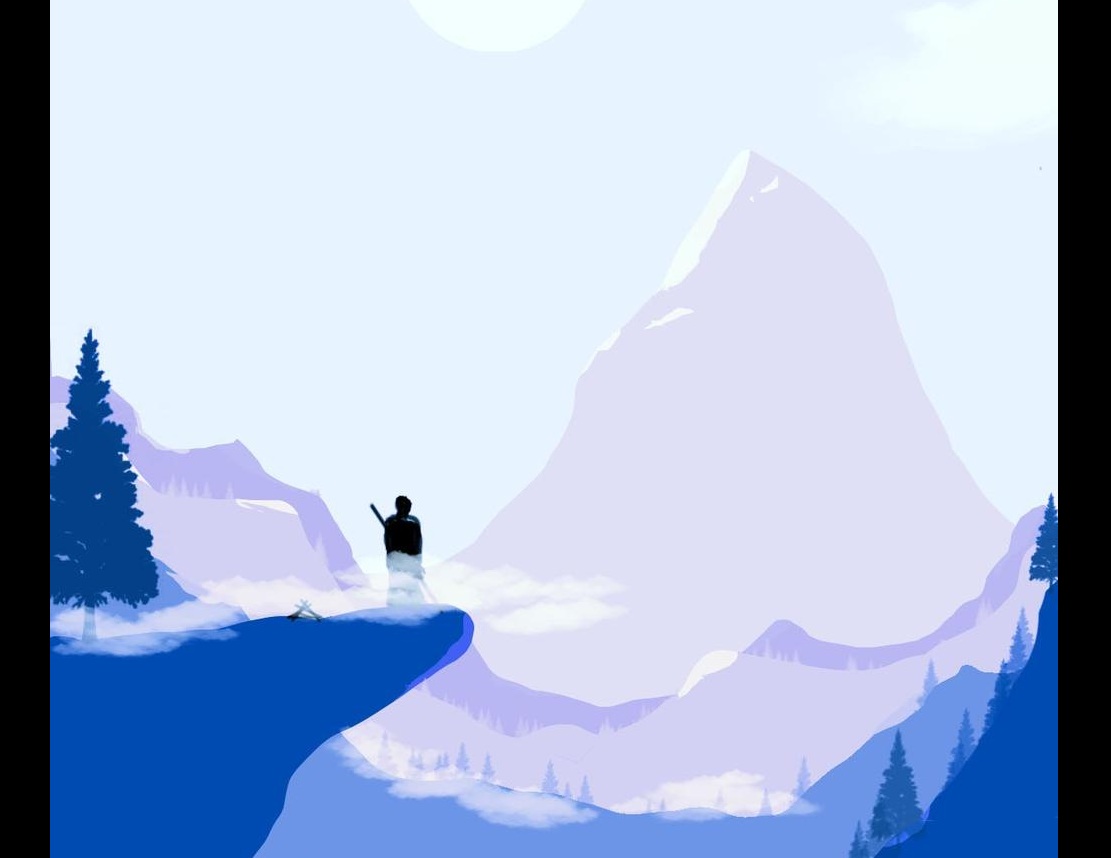}
            \vspace{-0.4em}

            {\scriptsize\raggedright hiker on snowy mountain..}
        \end{minipage}
        \hfill
        \begin{minipage}[t]{0.48\textwidth}
            \vspace{0pt}
            \centering
            \includegraphics[height=2.8cm, keepaspectratio]{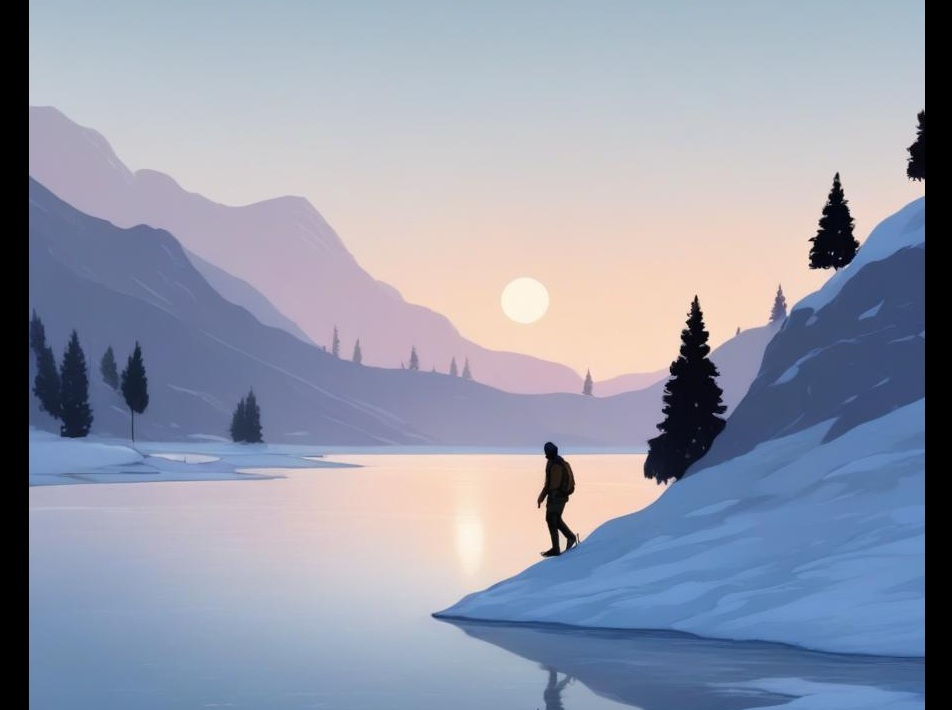}
            \vspace{-0.4em}

            {\scriptsize\raggedright walking down to a river..}
        \end{minipage}
        \caption{Change of scene-composition}
    \end{subfigure}

    \caption{Contextual variations in \textbf{original art}. Each pair shows original (left) and modified images (right).}
    \label{fig:contextual-variations-art}
\end{figure}

\section{Dataset Construction}

The primary contribution of Lunara Aesthetic II is its relational structure. Instead of treating images as independent samples, the dataset is organized into contextual anchor groups, where multiple variations are linked to a shared source image. This grouping makes the underlying supervision relational: each example is interpretable not only in isolation, but also in terms of what changed relative to the anchor and what should remain invariant. As a result, the dataset supports controlled analysis of contextual behavior that is difficult to obtain from standard i.i.d. collections.

All original images were obtained with explicit contributor consent. We first distilled initial image prompts using QWEN3-VL (\cite{qwen3technicalreport}), then generated 3,324 candidate variations with Moonworks Lunara, a diffusion mixture model that is trained using Moonworks CAT (Composite Active Transfer)\footnote{further details of Lunara and CAT will be provided in upcoming release}, a pretraining-time approach inspired by active learning~(\cite{hassan2025coherence,hassan-etal-2025-active,hassan-etal-2024-active,hassan-alikhani-2023-calm,Hassan2018interactive}).

The same VLM was used to identify contextual differences between each original–variation pair, followed by manual verification and refinement to ensure that visual contextual variations were accurately captured in the final labels. Finally, These variation outputs were manually filtered to retain 2.8K images exhibiting meaningful contextual changes.

For this contextual variation release, the goal is not to maximize visual diversity, but to construct \textit{identity-anchored} groups in which the underlying identity is preserved while targeted contextual attributes are systematically altered. This structure enables controlled evaluation of whether models learn contextual concepts that transfer across variations, rather than simply reproducing memorized imagery.

\section{Dataset Evaluations}

\begin{table}[]
\centering
\small
\begin{tabular}{lccc}
\toprule
contextual axis & Axis specificity & Prompt alignment (Cohen's $d$) & Axis entropy (n.) \\
\midrule
Illumination-time & 0.69 & -1.07 & 1.00 \\
Weather-atmosphere & 0.66 & -0.84 & 0.98 \\
Object-composition & 0.66 & -0.64 & 0.98 \\
Mood-atmosphere & 0.59 & -0.95 & 0.97 \\
Viewpoint-camera & 0.64 & -0.85 & 0.98 \\
Color-tone & 0.67 & -0.20 & 0.99 \\
\bottomrule
\end{tabular}
\caption{Automated evaluation of contextual variation axes. Axis specificity measures how well each axis is isolated from others, computed using conditional co-occurrence probabilities (higher values indicate greater isolation). Prompt alignment reports Cohen’s d for token-level similarity between prompts with and without a given axis, where more negative values correspond to larger prompt edits. Normalized entropy captures the diversity of realizations within each axis, with lower values indicating stronger axis isolation and higher values reflecting broader expressive variation.}
\label{tab:contextual_axis_eval}
\end{table}

\begin{table}[t]
\centering
\small
\begin{tabular}{lcccccc}
\toprule
\textbf{Dataset} & \textbf{N} & \textbf{Mean} & \textbf{Std} & \textbf{P05} & \textbf{P50} & \textbf{P95} \\
\midrule
Lunara-II-Variations & 2854 & \textbf{5.91} & 0.35 & \textbf{5.47} & \textbf{5.90} & \textbf{6.52} \\
Lunara-I & 2000 & \textbf{6.32} & 0.49 & \textbf{5.54} & \textbf{6.31} & \textbf{7.18} \\
CC3M & 1000 & 4.78 & 0.60 & 3.76 & 4.80 & 5.74 \\
LAION-2B-Aesthetic & 1000 & 5.25 & 0.44 & 4.55 & 5.25 & 5.95 \\
WIT & 1000 & 5.08 & 0.57 & 4.13 & 5.09 & 6.04 \\
\bottomrule
\end{tabular}
\caption{
Full distribution statistics of LAION Aesthetics v2 scores.
P05 and P95 denote the 5th and 95th percentiles, respectively.
}
\label{tab:aesthetic_distribution}
\end{table}

\subsection{Automated Evaluation}

We assess controlled contextual variation using prompt-level and distributional analyses.
Table~\ref{tab:contextual_axis_eval} reports automated metrics evaluating the isolation,
expressiveness, and prompt-level impact of each contextual variation axis.

Axis specificity values range from 0.59 to 0.69, indicating that all axes are reasonably
well isolated from one another under conditional co-occurrence analysis. Illumination–time
achieves the highest specificity (0.69), followed closely by color–tone (0.67) and
weather–atmosphere and object–composition (both 0.66). Mood–atmosphere exhibits the lowest
specificity (0.59), suggesting comparatively greater overlap with other contextual
dimensions, though still within a controlled range.

Prompt alignment, measured via Cohen’s $d$, shows consistently negative effect sizes across
all axes (-0.20 to -1.07), confirming that axis-conditioned variations induce meaningful
token-level prompt edits relative to their base prompts. Illumination–time exhibits the
largest divergence (-1.07), reflecting the explicit lexical changes required to modify
lighting and temporal conditions. Mood–atmosphere (-0.95), viewpoint–camera (-0.85), and
weather–atmosphere (-0.84) also show substantial prompt shifts, while color–tone produces
the smallest effect (-0.20), consistent with more localized lexical adjustments.

Normalized axis entropy values are uniformly high (0.97–1.00), indicating that each axis is
expressed through a broad and diverse set of prompt realizations rather than a small number
of repeated templates. Illumination–time reaches maximal entropy (1.00), while all other
axes remain close to this upper bound. In this setting, high entropy reflects lexical and
stylistic diversity within an axis rather than contextual leakage across axes, as
disentanglement is enforced through controlled prompt construction and labeling.

Table~\ref{tab:aesthetic_distribution} reports full distribution statistics of LAION
Aesthetics v2 scores across datasets. Lunara achieves the highest mean aesthetic score
(6.32) and the largest proportion of images exceeding the commonly used threshold of 6.5
(33.99\%), substantially outperforming CC3M, WIT, and LAION-2B-Aesthetic. Lunara-AIV exhibits
a lower mean score (5.91) and a smaller high-aesthetic tail (5.56\%), reflecting its focus
on controlled variation rather than aesthetic maximization. These results indicate that
controlled contextual variation does not inherently degrade perceptual quality and that
the Lunara dataset maintains strong aesthetic properties relative to widely used
benchmarks.

Overall, these automated metrics demonstrate that contextual variations are achieved
through substantive prompt-level interventions, maintain reasonable axis isolation, and
exhibit non-degenerate expressive diversity. Fine-grained instance-level consistency (e.g.,
unintended object introduction or structural drift) is not directly captured by these
metrics and is therefore addressed through targeted human evaluation.

\subsection{Aesthetic Preference Analysis}
We evaluate image aesthetics using the LAION Aesthetics v2 predictor, a CLIP-based model trained to approximate aggregate human judgments of visual appeal. We compare Lunara-AIV against several widely used vision–language datasets, including Lunara Art Aesthetic I (\cite{wang2026moonworkslunaraaestheticdataset}), Conceptual Captions (CC3M) (\cite{sharma2018conceptual}), a random subset of LAION-2B-Aesthetic (\cite{schuhmann2022laion}), and the Wikipedia-based Image–Text dataset (WIT) (\cite{srinivasan2021wit}). Table 3 reports full distributional statistics of predicted aesthetic scores.

Lunara-I exhibits the highest overall aesthetic quality, with the largest mean score (6.32) and a pronounced upper tail, as reflected by its high median (6.31) and 95th percentile (7.18). This distribution is consistent with strong aesthetic adherence during dataset construction, favoring visually striking and artistically composed imagery.

Lunara-II achieves a slightly lower mean aesthetic score (5.91) with substantially reduced variance, indicating a tighter but more moderate aesthetic distribution. This reduction relative to Lunara is intentional: Lunara-II explicitly includes everyday objects, routine environments, and visually ordinary scenes in order to support systematic evaluation of contextual variation. As a result, images are not optimized for visual appeal but instead reflect common real-world visual contexts, leading to relatively lower—but still high aesthetic scores.

In contrast, baseline datasets such as CC3M, LAION-2B-Aesthetic, and WIT exhibit lower mean scores (4.78–5.25) and weaker upper tails, reflecting more heterogeneous and less curated visual content drawn from large-scale web sources.

While Lunara prioritizes high-aesthetic imagery, Lunara-II occupies an intermediate regime: it maintains substantially higher aesthetic quality than common web-scale datasets while deliberately sacrificing some aesthetic peak performance to ensure broad coverage of everyday visual concepts and controlled contextual variation.

\subsection{Human Evaluation}
Independent researchers evaluate Lunara Aesthetic II at the level of an \textit{anchor-linked variation set} (an \textit{anchor image} and its associated contextual variants). Evaluation focuses on two dimensions of contextual consistency: \textbf{Identity Stability}, which measures whether identity is preserved across variants, and \textbf{Target Attribute Realization}, which measures whether the intended contextual change is clearly expressed.

\paragraph{Identity Stability.}
Researchers assess whether the core identity of the subject (e.g., the same object instance or entity) remains consistent across the anchor image and all variations. This criterion is necessary for attributing differences in model behavior to the intended contextual change rather than unintended identity drift. For each variation pair (an original image and a variation), evaluate using 5-point
Likert scale (1: poor stability, the variation does not retain original identity, 5: the variation image respects original identity).

\paragraph{Target Attribute Realization.}
Researchers assess whether the intended contextual factor changes clearly and meaningfully across the set (\textit{Yes/No}), conditioned on the set's target label. Each variation set is annotated with a target contextual factor describing the intended transformation. We consider six target variation types:
\begin{itemize}
    \item \textbf{Illumination-time} (e.g., day vs.\ night)
    \item \textbf{Weather-atmosphere} (e.g., snowy vs.\ cloudy)
    \item \textbf{Viewpoint\_camera} (e.g., neutral view vs.\ wide-angle shot)
    \item \textbf{Scene-composition} (e.g., indoor vs.\ outdoor; on top vs.\ under; broken vs.\ fixed)
    \item \textbf{Color-tone} (e.g., sepia; blue tone)
    \item \textbf{Mood-atmosphere} (e.g., soft, gentle, wild, windswept)
\end{itemize}

This design enables evaluation to be conditioned on the intended transformation type, rather than averaging across uncontrolled sources of variation.

\subsubsection{Human Evaluation results}

\begin{table}[t]
\centering
\begin{tabular}{c || c c c c c c}
\hline
\multicolumn{1}{c||}{\textbf{Identity}} & \multicolumn{6}{c}{\textbf{Target Attribute Realization (\% correct)}} \\
\cline{2-7}
\vspace{.2cm}
\textbf{Stability} & \textit{Illumination} & \textit{Weather} & \textit{Viewpoint} & \textit{Obj. Comp.} & \textit{Color} & \textit{Mood} \\
\hline
4.68 & 93.3 & 90.5 & 88.2 & 84.0 & 85.8 & 81.5 \\
\hline
\end{tabular}
\caption{Human evaluation results oon identity stability and target attribute realization}
\label{tab:human_eval_combined}
\end{table}

Results indicate that the generated image variations maintain a high degree of identity stability, with a mean Likert score of 4.68 across all evaluated samples. In terms of target attribute realization, performance is consistently strong across contextual dimensions, achieving 93.3\% accuracy for illumination changes and 90.5\% for weather-related variations. Viewpoint changes are correctly realized in 88.2\% of cases, while object composition modifications reach 84.0\% accuracy. More subtle appearance-related deltas show slightly lower but still robust performance, with 85.8\% accuracy for color tone and 81.5\% for mood or atmospheric changes. Overall, the results suggest that the model reliably implements a wide range of contextual transformations while preserving the core identity of the scene.

Evaluator feedback indicates that the model exhibits strong aesthetic quality and clear differentiation across visual themes, artistic styles, and atmospheric conditions, while generally maintaining high identity stability when the core subject is preserved. Variations in lighting and weather are rendered consistently and accurately, including subtle transitions such as changes in illumination time or atmospheric effects. However, evaluators noted that visually appealing variants can introduce bias in Likert-scale judgments of identity stability, potentially inflating perceived consistency.

Some unintended variability was also observed. In several cases, image composition changed substantially due to prompt phrasing (e.g., camera angle or scene description), and occasionally even in the absence of explicit instructions. Combinations such as wide-angle views paired with complex backgrounds sometimes produced overcrowded or unrealistic scenes, highlighting limits in compositional control under multi-factor variation.

The evaluation further revealed challenges in defining and applying semantic deltas. Several deltas—particularly color tone, illumination time, mood atmosphere, and weather atmosphere—were often intertwined and not strictly separable, leading to ambiguity in both model behavior and human assessment. Prompts that omitted explicit delta specifications raised additional questions about implicit model inference and its impact on evaluation outcomes. Evaluators recommended introducing clearer delta definitions, clustering overlapping deltas where appropriate, and providing explicit checklists for each variant. They also suggested expanding future evaluations to include greater subject-level variation, especially human-centric attributes such as pose or facial expression, to more rigorously assess identity stability beyond primarily scenic or inanimate content.

\section{Discussion}


Beyond its technical contributions, Lunara Aesthetic II is constructed with explicit attention to ethical data sourcing. All images are contributed by human creators with consent, reflecting a commitment to responsible dataset practices. Importantly, this choice is aligned with technical considerations: transparent provenance and intentional curation reduce noise and ambiguity in supervision signals, supporting more reliable semantic learning and evaluation. We view ethical data construction not as a trade-off with model capability, but as a practical foundation for developing generative systems that generalize reliably and behave predictably under semantic interventions.

Lunara Aesthetic II is intentionally released separately from the Lunara Aesthetics Dataset (\cite{wang2026moonworkslunaraaestheticdataset}). While the Aesthetics dataset targets stylistic conditioning and aesthetic control through broad prompt and artistic diversity, Lunara Aesthetic II focuses on contextual variation, controlled semantic transformation, and identity preservation under change. This separation clarifies intended use and evaluation: improvements on Lunara primarily reflect advances in style modeling, whereas improvements on Lunara Aesthetic II more directly measure contextual consistency and transformation behavior while maintaining high visual quality.

Lunara Aesthetic II supports three primary use cases: (1) evaluating generalization under controlled distribution shifts using held-out variations, (2) probing memorization and shortcut learning via identity stability across anchor-linked variants, and (3) benchmarking image editing and image-to-image systems where identity preservation under contextual transformation is critical while also preserving aesthetic quality. The dataset is designed for benchmarking, fine-tuning, and adaptation research, with evaluation naturally conducted both within-anchor consistency and across-anchor generalization.

We observe quality variance across anchor groups, which is primarily attributable to source-image characteristics rather than the variation procedure itself. In a small number of cases, the anchor images contain inherent ambiguity or reduced visual fidelity—such as low resolution, slight motion blur from handheld capture, or sketch-like/handwritten content with unclear boundaries. These factors can make the underlying identity less sharply defined and may lead to identity drift across variations. We retain these groups to reflect realistic acquisition conditions, but note that they may be less suitable for evaluations requiring strict pixel-level or fine-grained identity preservation.

The dataset’s moderate scale is also intentional. While larger datasets are essential for training frontier-scale models, they often limit interpretability and make it difficult to reason about what a model learned from any particular subset. Lunara Aesthetic II instead targets focused analysis and benchmarking, where the objective is not to saturate performance through scale, but to enable careful study of contextual behavior under controlled conditions. This makes the this release of dataset well suited for ablations, fine-tuning and adaptation studies, and evaluation of editing pipelines that must preserve identity while applying contextual changes.

Looking forward, Lunara-II provides a foundation for several extensions, including expanding contextual factor coverage, increasing variation depth per anchor, and developing standardized evaluation protocols that explicitly score identity consistency, edit specificity, and contextual correctness under controlled transformations. More broadly, relationally structured datasets of this form may help establish clearer empirical signals for when generative models genuinely acquire contextual concepts, rather than relying on superficial correlations.



\section{Conclusion and Future Release}
We introduced the Lunara Aesthetic II dataset, a public release of 2.8K relational contextual variations designed to support controlled analysis of contextual learning in modern image generation systems. By organizing images into identity-anchored variation sets with targeted contextual transformations, Lunara-II enables evaluation protocols that go beyond i.i.d.\ sampling, including the assessment of contextual generalization, memorization and shortcut learning, within-identity consistency, and robustness of image editing under controlled transformations. The dataset is released under the Apache 2.0 license to encourage broad reuse, reproducible benchmarking, and small-scale training and adaptation studies. 

In a future release, Moonworks will introduce the Lunara model, a novel diffusion mixture architecture trained with a new learning algorithm and used in the construction of both Lunara-I and Lunara-II.



\section*{Limitations and Ethical Considerations}

The dataset prioritizes controlled contextual analysis over scale and coverage. Variations are defined over a selected set of contextual transformations, reflecting explicit design choices about what constitutes meaningful change; consequently, more complex compositional, long-horizon, or highly abstract transformations are not exhaustively represented. It is not a substitute for broad pretraining corpora, but a structured benchmark for probing contextual learning and transformation robustness. In addition, identity preservation is defined operationally by the anchor--variation design, and downstream applications may require stricter or alternative notions of identity consistency.

From an ethical perspective, all images are derived from original, ethically sourced content and curated to avoid personally identifiable information (PII). We encourage responsible use aligned with the dataset's purpose: controlled evaluation of contextual transformations and identity-preserving edits.

\section*{Acknowledgment}
We thank Anthony Sicilia for his valuable feedback!

\bibliography{references}

\end{document}